\documentclass[11pt,a4paper,titlepage,doubleside]{memoir}  

\usepackage[OT1]{fontenc}

\usepackage[english]{babel}

\usepackage[utf8]{inputenc}

\usepackage[sc]{mathpazo}

\usepackage{amsmath,amssymb,amsfonts,mathrsfs}

\usepackage[amsmath,thmmarks]{ntheorem}

\usepackage{graphicx}

\usepackage{soul}

\usepackage{pdfpages}

\usepackage[format=plain,labelfont=bf,textfont=it]{caption}
\usepackage{subcaption}

\usepackage{natbib}
\setcitestyle{numbers,open={[},close={]},comma}
\usepackage{nicefrac}

\usepackage{bigints}

\usepackage[ruled, lined]{algorithm2e}

\usepackage{varioref}

\usepackage{datetime}

\usepackage{mathtools}

\usepackage[h]{esvect}

\usepackage{array}

\usepackage{listings}
\usepackage{booktabs}

\usepackage{ETHlogo}

\nonzeroparskip
\defaultlists

\makeatletter

\if@twoside
  \copypagestyle{chapter}{Ruled}
\else
  \copypagestyle{chapter}{ruled}
\fi
\makeoddhead{chapter}{}{}{}
\makeevenhead{chapter}{}{}{}
\makeheadrule{chapter}{\textwidth}{0pt}
\copypagestyle{abstract}{empty}

\makechapterstyle{bianchimod}{%
  \chapterstyle{default}
  \renewcommand*{\chapnamefont}{\normalfont\Large\sffamily}
  
  \renewcommand*{\printchaptername}{%
    \chapnamefont\centering\@chapapp}

  }

\chapterstyle{bianchimod}

\setsecheadstyle{\Large\bfseries\sffamily}
\setsubsecheadstyle{\large\bfseries\sffamily}
\setsubsubsecheadstyle{\bfseries\sffamily}
\setparaheadstyle{\normalsize\bfseries\sffamily}
\setsubparaheadstyle{\normalsize\itshape\sffamily}
\setsubparaindent{0pt}

\captionnamefont{\sffamily\bfseries\footnotesize}
\captiontitlefont{\sffamily\footnotesize}
\setsecnumdepth{subsection}
\settocdepth{subsection}

\pretitle{\vspace{0pt plus 0.7fill}\begin{center}\HUGE\sffamily\bfseries}
\posttitle{\end{center}\par}
\preauthor{\par\begin{center}\let\and\\\Large\sffamily}
\postauthor{\end{center}}
\predate{\par\begin{center}\Large\sffamily}
\postdate{\end{center}}

\def\@advisors{}
\newcommand{\advisors}[1]{\def\@advisors{#1}}
\def\@department{}
\newcommand{\department}[1]{\def\@department{#1}}
\def\@thesistype{}
\newcommand{\thesistype}[1]{\def\@thesistype{#1}}

\renewcommand{\maketitlehookb}{\vspace{1in}%
  \par\begin{center}\Large\sffamily\@thesistype\end{center}}

\renewcommand{\maketitlehookd}{%
  \vfill\par
  \begin{flushright}
    \sffamily
    \@advisors\par
    \@department, ETH Z\"urich
  \end{flushright}
}

\checkandfixthelayout

\makeatother

\theoremstyle{plain}
\numberwithin{equation}{chapter}

\theoremstyle{nonumberplain}
\theorembodyfont{\normalfont}
\theoremsymbol{\ensuremath{\square}}

\DeclareMathOperator{\rank}{rank}

\renewcommand{\epsilon}{\ensuremath\varepsilon}

\renewcommand{\phi}{\ensuremath{\varphi}}

\usepackage[linkcolor=black,colorlinks=true,citecolor=black,filecolor=black]{hyperref}

\title{Explicit Word Density Estimation for Language Modelling}
\author{Jovan Andonov}
\thesistype{Master Thesis}
\advisors{Advisors: Prof.\ Dr.\ T. Hofmann, PhD. O. Ganea, PhD. G. B\'{e}cigneul, PhD.\ P. Grnarova}
\department{Department of Computer Science}
\date{September 15, 2019}

\begin{document}

\frontmatter

\begin{titlingpage}
  \calccentering{\unitlength}
  \begin{adjustwidth*}{\unitlength-24pt}{-\unitlength-24pt}
    \maketitle
  \end{adjustwidth*}
\end{titlingpage}

\begin{abstract}
Language Modelling has been a central part of Natural Language Processing for a very long time and in the past few years LSTM-based language models have been the go-to method for commercial language modeling. Recently, it has been shown that when looking at language modelling from a matrix factorization point of view, the final Softmax layer limits the expressiveness of the model, by putting an upper bound on the rank of the resulting matrix. Additionally, a new family of neural networks based called NeuralODEs, has been introduced as a continuous alternative to Residual Networks. Moreover, it has been shown that there is a connection between these models and Normalizing Flows. In this work we propose a new family of language models based on NeuralODEs and the continuous analogue of Normalizing Flows and manage to improve on some of the baselines.
\end{abstract}

\newenvironment{acknowledgements}%
    {
    \begin{center}%
    \bfseries Acknowledgements\end{center}}%
    {\vfill\null}

\cleardoublepage
\begin{acknowledgements}

First and foremost, I would like to thank my family for always believing in me.

Next, I would like to thank my supervisors Octavian, Gary and Paulina for all the useful discussions. I would also like to thank the Data Analytics Lab and the Leonhard cluster for GPU access and free coffee.

A big thank you goes to Gorjan and Ondrej for all their advice, useful discussions and comments on the thesis text. I would also like to thank Igor for his constant support.

Last, but not least, I would like to thank Tina for the never-ending support and for always being there for me.

\end{acknowledgements}

\cleartorecto
\tableofcontents
\mainmatter

\chapter{Introduction}
\label{chapter:introduction}

Anything we see or do, can be described and contained within a sequence of words, meaning that the entire complexity of the world can be embedded in a piece of text. This is exactly what makes text and textual communication so important in our daily lives. Additionally, this is also what makes text processing and textual communication so complex for machines. Namely, the area of Computer Science that deals with these problems is called Natural Language Processing (NLP). NLP is a vast area with many subcategories, but without doubt, one of its core and most vital subcategories is text understanding and text generation.

In NLP, the tools that are used for text generation are called Language Models (LM). Let's consider the following sentence:

\begin{center}
    \emph{Look at all those clouds, it is going to ...}
\end{center}

Given the sentence above as a context, a Language Model will then try to estimate what is the most likely word to end the sentence. From a mathematical point of view, LMs try to learn a context conditioned probability distribution over a vocabulary. This means that given a vocabulary of available words and a sequence of words that represents the history or the context, a Language Model processes the context and returns a discrete probability distribution over the vocabulary

\begin{displaymath}
    P(w | w_{1..i-1}).
\end{displaymath}

Here $w_{1..i-1}$ is the context, often denoted as $ h $, and $ w $ is a discrete random variable that represents the vocabulary. We can then proceed with generating the next word by simply selecting the word with highest probability as

\begin{displaymath}
    \hat{w} = argmax_w P(w | w_{1..i-1})
\end{displaymath}

However, very often we do not want to only generate a single word given a context, but instead we want to generate whole sequences. Therefore, the LMs can also be seen as tools that model the joint probability distribution over a textual sequence. Or mathematically speaking

\begin{displaymath}
    P(w_1, ..., w_n) = \prod_i^n P(w_i | w_{1..i-1}).
\end{displaymath}

First LMs were count based and called N-grams \citep{martin2009speech}. However, with the recent advances in Deep Learning, LMs based on Neural Networks are currently dominating the field. Since the first Neural Language Model \citep{bengio2003neural} which was based on Feedforward Neural Networks, things have evolved and now Recurrent Neural Networks (RNN) \citep{mikolov2010recurrent} are the standard. Additionally, as neural networks are trained with gradient based methods and back-propagation \citep{rumelhart1988learning}, people have figured out that RNNs, when processing long contexts can suffer from the vanishing or exploding gradients problem \citep{hochreiter1998vanishing, pascanu2012understanding, pascanu2013difficulty}. Therefore, Vanilla RNNs were substituted with Long short-term Memory (LSTM) \citep{hochreiter1997long} based RNNs. LSTMs alleviate the vanishing gradient problem and additionally gradient clipping \citep{pascanu2013difficulty} takes care of the exploding gradients problem. Recently, Transformer \citep{vaswani2017attention} based models like BERT \citep{devlin2018bert} and GPT-2 \citep{radford2019language} have enjoyed quite the success in language modelling and language understanding tasks. However, these models have an enormous number of parameters and need an enormous amount of resources to be trained. Therefore, in the past few years, LSTM-based RNNs with a Softmax layer on top have been the go-to method for commercial language modelling.

This thesis first gives an overview of the current limitations of Language Modelling. Then it describes how previous work has tried to break these limitations. Then, introduces a novel idea for overcoming the previously mentioned limitations of Language Modelling. Towards the end, it describes the architecture of the models created in the scope of this thesis, as well as present the results from the experiments. Finally, it concludes the findings and suggests possible future work.

\chapter{Current Limitations of Language Models}
\label{chapter:limitations}

\section{The Softmax Bottleneck}
\label{section:limitations:softmax_bottleneck}

The majority of parametric LMs use a Softmax function operating on the context $ c $, and a word embedding $ e_w $ to define the conditional distribution $ P_\theta(w|c) $, where $ w $ stands for a word, and $ \theta $ are the parameters of the model. More specifically, the model distribution is usually written as

\begin{displaymath}
    P_\theta(w | c) = \frac{h^Te_w}{\sum_{e_{w'}} h^Te_{w'}},
\end{displaymath}

where $ h $ is a function of $ c $ and it is commonly obtained using a Recurrent Neural Network, and $ e_w $ is the embedding for word $ w $. Additionally, $ h \in R^D $ and $ e_w \in R^D $ and both of them depend on $ \theta $. Finally, we refer to the dot product $ h^Te_w $ as a \emph{logit}.

Even though, one might argue that natural languages contain an infinite amount of contexts, let's take a look at the finite case first. We can assume that a natural language consists of \emph{N} contexts and \emph{M} words. Consequently, we can describe Language Modelling as a matrix factorization problem. Consider the following matrices:

\begin{displaymath}
    \begin{matrix}
        H_\theta &= \begin{bmatrix}
               h^T_{c_1} \\
               h^T_{c_2} \\
               \vdots \\
               h^T_{c_N}
              \end{bmatrix}
        &     
        E_\theta &= \begin{bmatrix}
           e^T_{w_1} \\
           e^T_{w_2} \\
           \vdots \\
           e^T_{w_M}
          \end{bmatrix}
    \end{matrix} \\
\end{displaymath}
\begin{displaymath}
    \begin{matrix}
    A &= \begin{bmatrix}
       \log P(w_1 | c_1) & \log P(w_2 | c_1) & \hdots & \log P(w_M | c_1)  \\
       \log P(w_1 | c_2) & \log P(w_2 | c_2) & \hdots & \log P(w_M | c_2) \\
       \vdots & \vdots & \ddots & \vdots \\
       \log P(w_1 | c_N) & \log P(w_2 | c_N) & \hdots & \log P(w_M | c_N)
      \end{bmatrix}
    \end{matrix}
\end{displaymath}

Where $ H_\theta \in R^{N \times D} $ is a matrix containing the hidden states for every context as row vectors. $ E_\theta \in R^{M \times D}$ is a matrix containing word embeddings for every word as row vectors. $ A $ is a matrix containing the true log probabilities of every word, given every context. Then, language modelling can be described as:

\begin{displaymath}
    H_\theta E^T_\theta = \hat A
\end{displaymath}

Where, $ \hat A $ is:

\begin{displaymath}
    \begin{matrix}
    \hat A &= \begin{bmatrix}
       \log P_\theta(w_1 | c_1) & \log P_\theta(w_2 | c_1) & \hdots & \log P_\theta(w_M | c_1)  \\
       \log P_\theta(w_1 | c_2) & \log P_\theta(w_2 | c_2) & \hdots & \log P_\theta(w_M | c_2) \\
       \vdots & \vdots & \ddots & \vdots \\
       \log P_\theta(w_1 | c_N) & \log P_\theta(w_2 | c_N) & \hdots & \log P_\theta(w_M | c_N)
      \end{bmatrix}
    \end{matrix}
\end{displaymath}

and we want it to be as close as possible to the true $ A $. Now we can ask the following question:

\begin{center}
    \emph{"What is the expressiveness of this language model?"}
\end{center}

We can then proceed to answer this question from a matrix factorization point of view. Essentially, we want to learn matrices $ H_\theta $ and $ E_\theta $ such that we will be able to factorize the true distribution $ A $. However, in order for a valid factorization to exist the rank of $ H_\theta E^T_\theta $ has to be at least as large as the rank of $ A $, i.e. $ \rank(H_\theta E^T_\theta ) \geq \rank(A) $. As $ H_\theta \in R^{N \times D} $ and $ E_\theta \in R^{M \times D} $, $ \rank(H_\theta E^T_\theta ) $ is bounded by $ D $. Therefore, this is a limitation that comes from the final \emph{Softmax layer}. It simply means, that no matter how efficient we are in embedding all contexts into a matrix $ H_\theta $, we will not be able to retrieve the true language distribution $ A $, unless $ D \geq \rank(A) $.

To realize why this is indeed a bottleneck, and a problem in language modelling, we should first consider the typical dimensionalities that are used for the hidden state and the word embeddings. Usually, $ D $ is in the low hundreds, while the rank of the true distribution $ A $ can theoretically be up to $ M $ which is usually at least as large as $ 10^4 $. Right off the bat, we have a mismatch of several orders of magnitudes. One might say that an easy fix is to simply increase $ D $ and have a $ M \times M $ \emph{Softmax} in the final layer. However, this will drastically increase the amount of trainable parameters, resulting in slower training and harder optimization. Even though, wider and larger neural networks are theoretically more expressive, in practice they are a lot more difficult to train.

As using a larger $ D $ is not a straightforward solution to the problem it means that typical language models are \emph{Low Rank Language Models}. This would only cause problems if the true distribution $ A $ is indeed of a high rank. It is very hard to prove that natural languages are of high rank. However, intuitively speaking, if the true distribution of a natural language was indeed to be of a low rank, it would mean that all semantic meanings can be created by combining a small number of meanings. Which seems very odd and no linguist has ever managed to find such a small subset of bases, which can fully describe a language. Therefore, \citet{yang2017breaking} speculate that a high rank language model is needed to capture the true distribution.

\section{Single Transformation}
\label{section:limitations:single_transformation}

Another limitation of classical language models is the output projection layer. Taking the matrix-vector product between the embedding matrix (nowadays weights between the output projection and the embedding matrix are usually tied) and the hidden state, further exponentiating and normalizing the result to get a probability distribution, essentially results in a transformation that behaves as a single mode Gaussian around the hidden state. This means that the majority of the probability mass is concentrated on one very small continuous subspace of the embedding space. This subspace usually corresponds to the surroundings of the word embedding that is mostly associated with the given context throughout training. Take the following sentence as an example:

\begin{center}
    \emph{“I want to buy…”}
\end{center}

\begin{figure}
\centering
\begin{subfigure}{.5\textwidth}
  \centering
  \includegraphics[width=\linewidth]{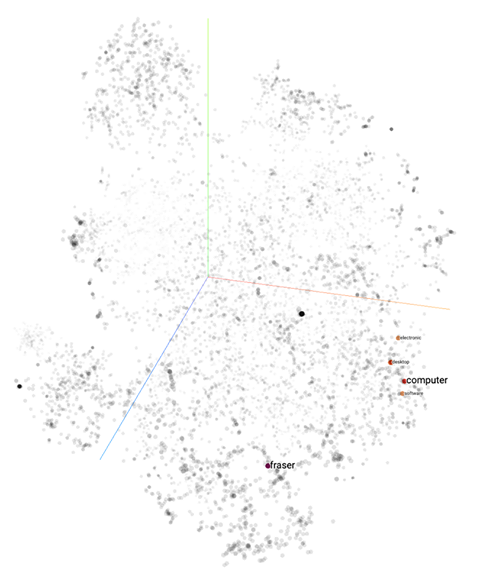}
  \label{figure:car:neighborhood}
\end{subfigure}%
\begin{subfigure}{.5\textwidth}
  \centering
  \includegraphics[width=\linewidth]{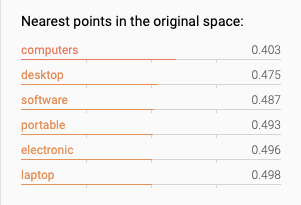}
  \label{figure:car:neighbor}
\end{subfigure}
\caption{Neighborhood of \emph{car} in the embedding space of AWD-LSTM.}
\label{figure:car}
\end{figure}

\begin{figure}
\centering
\begin{subfigure}{.5\textwidth}
  \centering
  \includegraphics[width=\linewidth]{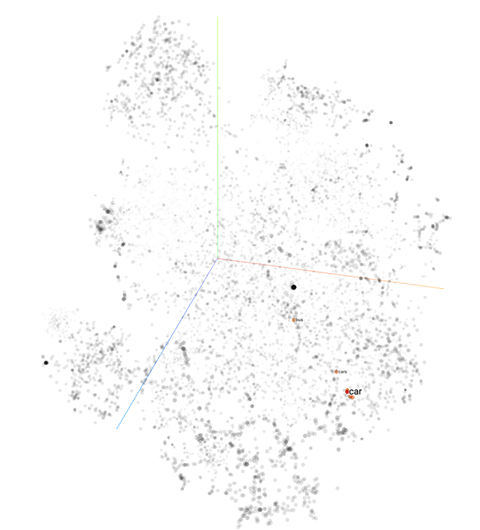}
  \label{figure:computer:neighborhood}
\end{subfigure}%
\begin{subfigure}{.5\textwidth}
  \centering
  \includegraphics[width=\linewidth]{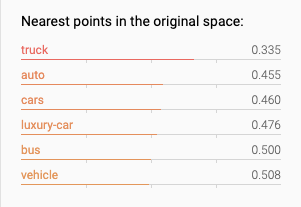}
  \label{figure:computer:neighbor}
\end{subfigure}
\caption{Neighborhood of \emph{computer} in the embedding space of AWD-LSTM.}
\label{figure:computer}
\end{figure}

It is easy to see that multiple words are likely to come after this particular context even though they might be associated with different contexts on average. This is a very common scenario in natural languages where the distribution over the vocabulary is considered to be a \emph{fat tail} distribution. For example take the words \emph{car} and \emph{computer}. By training a Language Model and plotting the embeddings, it is possible to visualize that the neighbors of \emph{car} in the embedding space are \emph{truck, auto, bus, vehicle} etc. and the neighbors of \emph{computer} are \emph{desktop, portable, electronic, laptop} etc. Refer to figures \ref{figure:car} and \ref{figure:computer}. This indicates that \emph{car} and \emph{computer} are in different neighborhoods of the embedding space.

Now imagine every single word that can finish the previous sentence. Every one of those words is embedded into its own neighborhood where it is close to other words that share the same context on average. Therefore, a simple single mode Gaussian seems very limited for all those \emph{``fat tail''} situations in natural languages. Ideally, we would like something that can adapt based on the context, i.e. given a hidden state we can obtain different type of distributions. It is also possible that the success of MoS \citep{yang2017breaking} and DOC \citep{takase2018direct} papers, in addition to breaking the \emph{softmax bottleneck}, is due to the fact that they learn a dynamic mixture of Gaussians, which by definition is more expressive. Additionally, based on the amount of Gaussians in the mixture, those approaches are more or less capable of solving the previously mentioned issues.

To generalize, it seems unlikely that a single general distribution, conditioned on the hidden state, can capture all possible situations in Language Modelling. Therefore, it seems reasonable that based on the context, or the hidden state if you want, we would like to obtain custom distributions that specifically fit the need of the current context. But what if we can start with a ``simple'', single, general distribution and then based on the context, distort it into a more complex distribution if needed? That would alleviate the need to manually tune the amount of components in the mixture and can be seen as a generalization of MoS \citep{yang2017breaking} and DOC \citep{takase2018direct}.

\chapter{Related Work}
\label{chapter:related_work}

\section{AWD-LSTM}
\label{section:related_work:awd_lstm}
\citet{merity2017regularizing} had several crucial contributions to RNN-based language modelling with their AWD-LSTM model.

First, regularizing RNNs is a complicated matter. A na\"ive application of dropout \citep{srivastava2014dropout}, where we randomly dropout units in every time step, results in unit starvation and disrupts the RNN's ability to retain long term dependencies. \citet{gal2016theoretically} suggest using the same dropout mask in every time step, to prevent this from happening. On the other hand, \citet{merity2017regularizing} propose the weight-dropped LSTM, which uses Dropconnect \citep{wan2013regularization} on the hidden-to-hidden weights as a recurrent regularization.

Secondly, they propose the use of Non-monotonically Triggered Averaged Stochastic Gradient Descent (NT-ASGD) as an optimization algorithm, instead of the regular Stochastic Gradient Descent (SGD).

The training of neural networks can be defined as

\begin{displaymath}
    \hat{\theta} = \underset{\theta}{argmin} \ \frac{1}{N} \sum^N_{i=1} f_i(\theta),
\end{displaymath}

where $ f_i $ is the loss function for the $ i $-th data point and $ \theta $ are the parameters to be learned. SGD takes the form of

\begin{displaymath}
    \theta_{k+1} = \theta_k + \gamma_k \hat{\nabla} f(\theta_k),
\end{displaymath}

where $ k $ stands for the iteration number, $ \gamma_k $ is the learning rate and $ \hat{\nabla} $ denotes a stochastic gradient on a minibatch of samples. After convergence, SGD returns the final iteration as the solution. Contrary to this, Averaged SGD (ASGD) returns the average

\begin{displaymath}
    \frac{1}{K-T+1} \sum_{i=T}^K \theta_i
\end{displaymath}

as a solution. Here $ K $ is the total number of iterations and $ T < K $ is a user-specified averaging trigger. \citet{merity2017regularizing} propose to use an automatic trigger mechanism for the averaging. Instead of manually specifying a value for $ T $, they propose a non-monotonic criterion that triggers the averaging when the validation metric does not improve for several epochs. The full method is shown in algorithm \ref{algorithm:related_work:ASGD}.

\begin{algorithm}[H]
\SetAlgoLined
\SetKwInOut{Input}{Inputs}
\Input{Initial parameters $ \theta_0 $, learning rate $ \gamma $, logging interval $ L $, non-monotone interval $ n $}
 Initialize $k \leftarrow 0, \ t \leftarrow 0, \ T \leftarrow 0, \ logs \leftarrow [] $ \\
 \While{stopping criterion not met}{
  Compute stochastic gradient $ \hat{\nabla}f(\theta_k) $ and take the SGD step \\
  \If{mod(k, L) = 0 and T = 0}{
   Compute validation perplexity $ v $ \\
   \If{t $ > $ n and v $ > $ \ $ \underset{l \ \in \ {t-n, \ ... , \ t}}{min} \ logs[l] $ }{
    Set $ T \leftarrow K $
   }
   Append $ v $ to $ logs $ \\
   $ t \leftarrow t + 1 $
   }
 }
 \Return $ \frac{1}{k - T + 1} \sum_{i=T}^k \theta_i $
 \caption{Non-monotonically Triggered ASGD (NT-ASGD) \citep{merity2017regularizing}\label{algorithm:related_work:ASGD}}
\end{algorithm}

Finally, their codebase \footnote{https://github.com/salesforce/awd-lstm-lm} set a new foundation for RNN-based language modelling. Every other notable model that came out in the next few years used their codebase as a basis to build upon.

\section{Mixture of Softmaxes}
\label{section:related_work:mos}

\citet{yang2017breaking} contributions in their paper \emph{Breaking the Softmax Bottleneck} are two-fold. First, they identify the Softmax Bottleneck problem described in section \ref{section:limitations:softmax_bottleneck} and secondly, they propose a simple technique that allows to bypass this limitation.

They use the AWD-LSTM model to encode the context in a vector $ g_c $. Then, they project $ g_c $ to obtain multiple hidden states

\begin{displaymath}
    h_{c, k} = tanh(W_{h,k} g_c)
\end{displaymath}

where $ h_{c, k} $ stands for the $ k $-th component for context $ c $ and $ W_k $ are trainable parameters. Then, the conditional probability of a word given a context is modelled as

\begin{align*}
    P_\theta(w | c) = &\sum_{k=1}^K \pi_{c, k} \frac{\exp{h^T_{c, k} e_w}}{\sum_{w'} \exp{h^T_{c, k} e_{w'}}} \\
    s.t. &\sum_{k=1}^K \pi_{c, k} = 1,
\end{align*}

where $ e_w $ is the (output) word embedding for word $ w $. The prior weights $ \pi_{c, k} $ are obtained by

\begin{displaymath}
    \pi_{c, k} = \frac{\exp{w_{\pi, k}^T g_c}}{\sum_{k'=1}^K \exp{w_{\pi, k'}^T g_c}}.
\end{displaymath}

According the previous equations, we can deduce that \citet{yang2017breaking} propose to have a weighted average between several Softmax functions. Therefore, they call this model \emph{Mixture of Softmaxes (MoS)}. Furthermore, if we create a matrix $ \hat{A}_{MoS} $ similar to the one in section \ref{section:limitations:softmax_bottleneck}, we get

\begin{displaymath}
    \hat{A}_{MoS} = \log \sum_{k=1}^K \Pi_k \exp (H_{k}E).
\end{displaymath}

As $ \hat{A}_{MoS} $ is now obtained via a non-linear transformation, we can deduce that its rank is not bounded and $ \hat{A}_{MoS} $ can potentially be a full-rank matrix.

\section{Direct Output Connections}
\label{section:related_work:doc}

\citet{takase2018direct} propose their \emph{Direct Output Connections (DOC)} model as a generalization of MoS. DOC computes $ J $ probability distributions from all layers and performs a weighted average between them. The output probabilities in DOC, are computed as

\begin{align*}
    P_\theta(w | c) = &\sum_{j=1}^J \pi_{j, c} \ Softmax(\Tilde{W}k_{j,c}) \\
    s.t. &\sum_{j=1}^J \pi_{j, c} = 1
\end{align*}

where $ \pi_{j, c} $ is the weight for the $ j $-th component in the mixture given context $ c $ and is obtained by

\begin{displaymath}
    \pi_{c} = Softmax(W_\pi h^N_c),
\end{displaymath}

where $ \pi_{c} $ is a vector with elements $ \pi_{j, c} $, $ W_\pi $ is a weight matrix and $ h^N $ is the hidden state from the final layer. Furthermore, $ \Tilde{W} \in R^{|V| \times d} $ is a weight matrix and $ k_{j,c} \in R^d $ is a vector computed from the hidden state of some layer $ n $ as

\begin{displaymath}
    k_{j, c} = W_j h^n_c.
\end{displaymath}

In this equation $ W_j \in R^{d \times d_{h^n}} $ is a weight matrix. Additionally, let $ i_n $ be the number of $ k $-s computed from the hidden state of the $ n $-th layer s.t. $ \sum_{n=0}^N i_n = J $. From here we can deduce that for $ i_N = J $, i.e. if all distributions are obtained from the final layer, DOC is equivalent to MoS, which is exactly why DOC is considered to be a generalization of MoS.

Finally, if we construct the matrix containing all log-probabilities given all contexts for DOC, we can notice that it takes the form of

\begin{displaymath}
    \hat{A}_{DOC} = \log \sum_{j=1}^J \Pi \ Softmax(K_j \Tilde{W}^T)
\end{displaymath}

where $ \Pi $ is a diagonal matrix whose entries are the weights $ \pi_{j, c} $ and $ K_j $ is a matrix whose rows are vectors $ k_{j,c} $. As $ \hat{A}_{DOC} $ is obtained using a non-linear transformation, $ \hat{A}_{DOC} $ can be of an arbitrary high rank and is not limited by the Softmax Bottleneck explained in section \ref{section:limitations:softmax_bottleneck}.

\chapter{Neural Ordinary Differential Equations}
\label{chapter:ode}

\section{Introduction to Neural ODEs}
\label{section:ode:introduction}

In recent years Residual Networks (ResNet) \citep{he2016deep} have brought a great success in Deep Learning and especially in computer vision. They have proven to be effective against the vanishing gradient and the degradation problems and have drastically eased the optimization of very deep neural networks. If we refer to the output vector of each layer as $ z_t $ where $ t $ stands for the layer, then Residual Networks can be mathematically described as

\begin{equation}
    \label{equation:ode:resnet}
    z_{t+1} = z_{t} + f(z_{t}; \ \theta_{t}),
\end{equation}

where $ t \in \{0, ..., T\} $, $ z_t \in R^D $ and $ \theta_t $ are the parameters of the \emph{t}-th layer. These iterative updates can be interpreted as an Euler discretization of a continuous transformation \citep{lu2017beyond, haber2017stable, ruthotto2018deep}.

Moreover, as we add more layers and take smaller steps, in the limit, we parameterize the continuous dynamics of the hidden state using an ordinary differential equation (ODE) specified by a neural network

\begin{equation}
    \label{equation:ode:odes}
    \frac{d z(t)}{d t} = f(z(t), \ t; \ \theta ).
\end{equation}

\begin{figure}[ht]
      \centering
      \includegraphics[width=0.7\columnwidth]{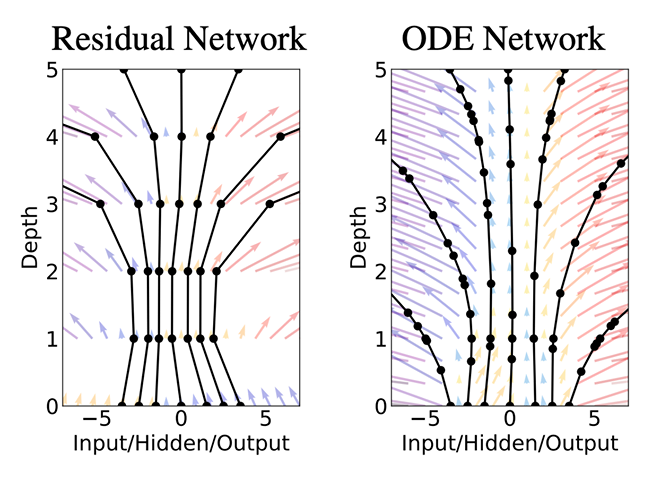}
      \caption{\emph{Left}: A Residual network defines a discrete sequence of finite transformations. \emph{Right}: An ODE network defines a vector field that continuously transforms the state \citep{chen2018neural}.}
      \label{figure:ode:resnet_vs_ode}
\end{figure}

\citet{chen2018neural} introduced this concept as a new family of deep neural network models, where the neural network outputs the gradient of the hidden state with respect to the depth. Then, given an initial state and the differential equation parameterized by the neural network, the final state is obtained by solving an ODE. The analogy they make is the one that considers this family of models to be the continuous case of ResNets. Figure \ref{figure:ode:resnet_vs_ode}, depicts the similarities and differences between ResNets and neural networks based on ODEs. They call this family of models \emph{Neural ODEs} or \emph{ODENets}, and provide an open source framework implemented in PyTorch\footnote{\url{https://github.com/rtqichen/torchdiffeq}}.

In equation \ref{equation:ode:odes}, $ z(t) $ is the hidden state in the \emph{t}-th layer, and $ f $ can be any neural network parameterized by $ \theta $, with $ z(t) $ and $ t $ as inputs and the gradient of $ z(t) $ with respect to $ t $ as output. Furthermore, the final output of such a model can then be defined as

\begin{align}
    z(t_1) & = z(t_0) + \int_{t_0}^{t_1} \frac{d z(t)}{d t} \ dt \\
    & = z(t_0) + \int_{t_0}^{t_1} f(z(t), \ t; \ \theta ) \ dt \\
    & = ODESolve(z(t_0), f, t_0, t_1).
\end{align}

\begin{figure}[ht]
      \centering
      \includegraphics[width=\columnwidth]{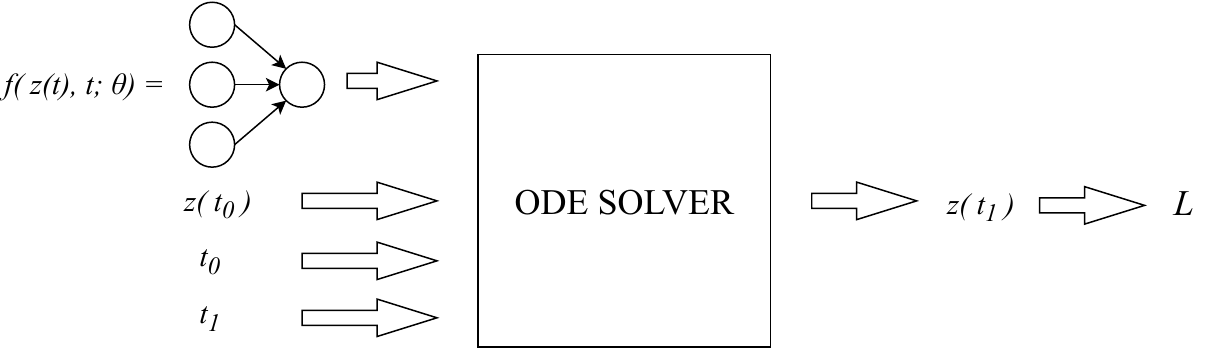}
      \caption{General purpose ODENet. $f(z(t), \ t; \ \theta) $ is neural network specifying the differential equation, $ z(t_0) $ is the initial state, $ t_0 $ is the initial time, $ t_1 $ is the final time, $ z(t_1) $ is the final state and $ L $ is a scalar valued loss function.}
      \label{figure:ode:odenets_visualization}
\end{figure}

According to the equations above, we can conclude that, $ f $ is learning a vector field, which is why Neural ODEs can potentially be seen as models with infinite amount of layers. To be specific, the number of layers is dynamically decided and delegated to the ODE solver. Furthermore, \citet{chen2018neural} developed their framework in way that any ODE solver can be used as a blackbox. This allows for more flexibility and decouples the framework from the ODE solver. General purpose ODENets are illustrated on Figure \ref{figure:ode:odenets_visualization}.

Defining neural network models in this fashion has several advantages \cite{chen2018neural}:
\begin{itemize}
    \item \textbf{Memory Efficiency.} In section \ref{section:ode:backpropagation} it is discussed how circumventing backpropagation through the operations of the ODE solver saves a lot of memory.
    \item \textbf{Adaptive Computation.} Nowadays, ODE solvers provide guarantees about the growth of the approximation error, monitor the error, and adapt their evaluation strategy on the fly to achieve the required level of accuracy. This allows for explicit control over the speed versus precision trade-off.
    \item \textbf{Parameter Efficiency.} In section 3 of their work, \citet{chen2018neural} demonstrate how this family of models ties the weights of nearby layers, resulting in fewer parameters without the loss of performance.
    \item \textbf{Scalable and invertible normalizing flows.} Chapter \ref{chapter:cnf} sshows how one side effect of going in the continuous domain, allows for an easier and unrestricted use of normalizing flows. As a result, normalizing flows can be used for language modelling, as shown in chapter \ref{chapter:cnf_lm}.
    \item \textbf{Continuous time-series models.} RNNs are the de-facto architecture for time-series models. Unfortunately, they require that the observations are discretized and bound to specific emission intervals. On the other hand, continuously defined dynamics can naturally take care of observations that arrive at arbitrary times.
\end{itemize}

\section{Backpropagation through the ODE solver}
\label{section:ode:backpropagation}

An immediate question that rises, is how does one backpropagate through the ODE solver. In theory one can simply backpropagate through the operations of the solver, however, this has several drawbacks. First, some solvers, require solving a nonlinear optimization problem at every step. This can make direct backpropagation through the integrator difficult. Additionally, as mentioned in the previous section, ODENets can potentially have a very high number of layers. Backpropagating through such a large number of layers is inefficient from a memory point of view, as it would mean that all intermediate steps should be kept in memory until the backward pass is over. Therefore, what \citet{chen2018neural} propose, is to compute gradients by solving a second augmented ODE backwards in time. This method is called the \emph{adjoint sensitivity method} \citep{pontryagin2018mathematical} and is applicable to all ODE solvers. It scales linearly with problem size, has low memory cost, and allows for explicit control over numerical errors.

\begin{figure}[ht]
      \centering
      \includegraphics[width=\columnwidth]{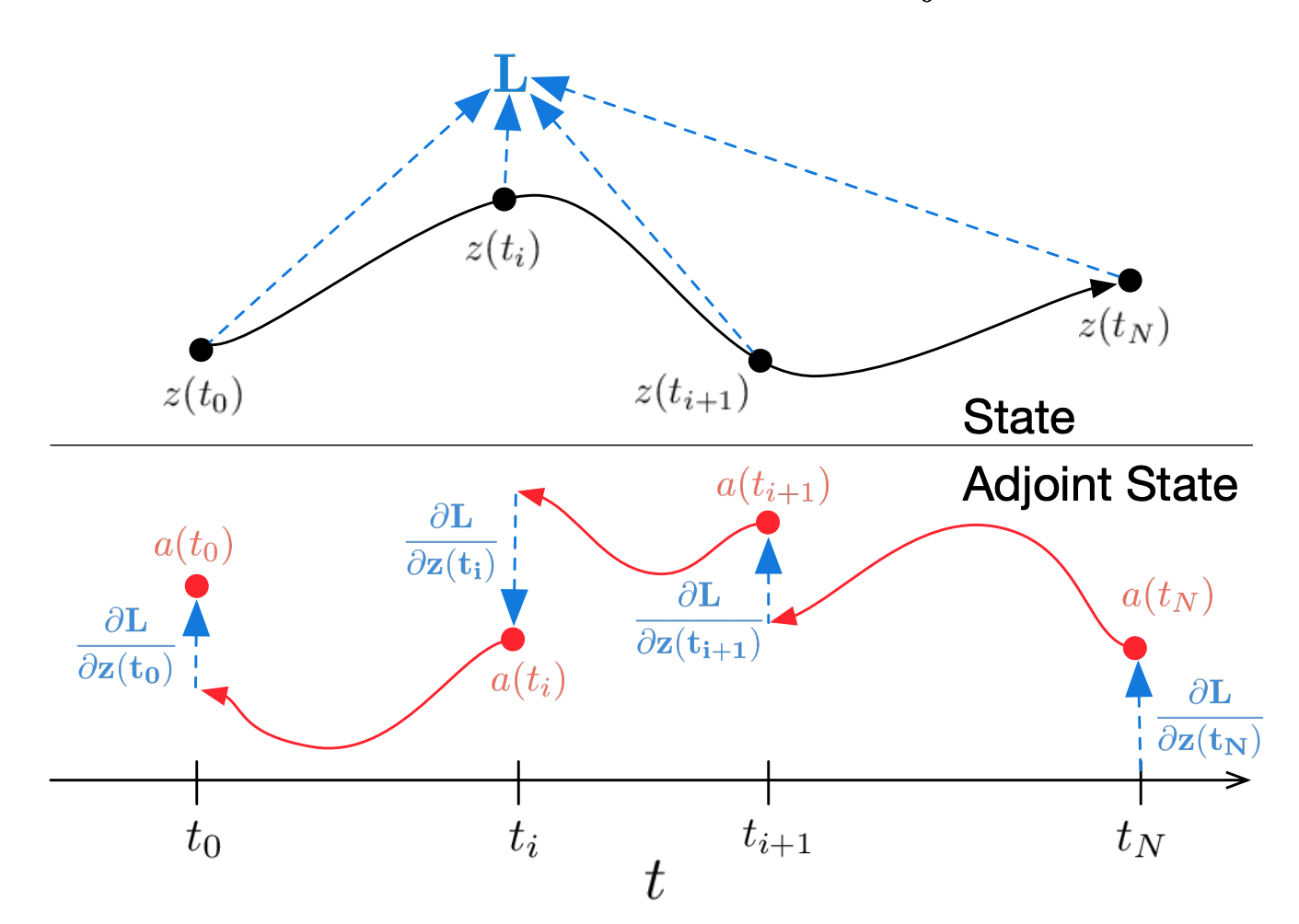}
      \caption{Reverse-mode differentiation of an ODE solution. The adjoint sensitivity method solves an augmented ODE backwards in time. The red lines denote the sequence of separate ODE solves \cite{chen2018neural}.}
      \label{ode:adjoint}
\end{figure}

As shown in figure \ref{figure:ode:odenets_visualization}, we need to optimize a scalar valued loss function $ L $ whose input is the output of the ODE Solver.

\begin{equation}
    L(z(t_1)) = L \left( z(t_0) + \int_{t_0}^{t_1} f(z(t), \ t; \ \theta ) \right) = L(ODESolve(z(t_0), f, t_0, t_1))
\end{equation}

The parameters of the ODENet are the parameters of the neural network $ f $, which is why we are interested in the derivative of the loss function with respect to $ \theta $. \citet{pontryagin2018mathematical} show that the derivative takes the form of another initial value problem

\begin{equation}
    \label{loss_derivative}
    \frac{dL}{d\theta} = - \int_{t_1}^{t_0} \left(\frac{\partial L}{\partial z(t)}\right)^T \frac{\partial f(z(t), \ t; \ \theta)}{\partial \theta}dt
\end{equation}

Where $ \nicefrac{\partial L}{\partial z(t)} $ is known as the \emph{adjoint state} of the ODE. \citet{chen2018neural} use one call to an ODE solver to get $ z(t_1) $ and then a second one to calculate the equation \ref{loss_derivative}. In cases where the loss depends not only on the final state $ z(t_1) $, but also on the intermediate states $ z(t) $, the reverse-mode derivative must be broken into a sequence of separate solves as shown on figure \ref{ode:adjoint}.

\section{Neural ODEs for Language Modelling}
\label{section:ode:ode_lms}

We can look at language modelling as a two step process. First, we encode the context into a hidden state vector, and then using the hidden state vector we generate a distribution over the vocabulary. The former is commonly done by an LSTM-based RNN. Then, given the hidden state, a Softmax layer is used to obtain the probability distribution over the vocabulary. In section \ref{section:limitations:softmax_bottleneck}, it is discussed how having this Softmax layer introduces a theoretical limitation on LMs, and in sections \ref{section:related_work:mos} and \ref{section:related_work:doc} is discussed how models like MoS \citep{yang2017breaking} and DOC \citep{takase2018direct} have tried to break it. In this section, a simple model based on Neural ODEs is proposed, that is not restricted by the Softmax Bottleneck problem.

Let $ V $ be the vocabulary, $ h $ be the hidden state, $ d $ be the hidden state dimensionality, $ L \in R^{d \times |V|} $ be the output projection matrix (if weights are tied \citep{inan2016tying} this is also the embedding matrix), and $ y^* \in R^{|V|} $ be a one-hot encoded ground truth vector. The model can then be represented as

\begin{align*}
    l(t_0) &= L^T h, \ l \in R^{|V|} \\
    l(t_1) &= node(l(t_0), \ f) \\
    y &= Softmax(l(t_1)),
\end{align*}

where $ node $ is a NeuralODE block defined as

\begin{displaymath}
    node(l(t_0), \ f) = l(t_0) + \int_{t_0}^{t_1} f(l(t), \ t) dt    
\end{displaymath}

and $ f $ is a neural network parameterizing the gradient of the logits $ l $ with respect to time or in this case depth. Moreover, $ f $ can be an arbitrary neural network architecture and one possibility is to define it as

\begin{displaymath}
    f(l, t) = H_f ReLU(W_f^T l), \ W_f, H_f \in R^{|V| \times |V|}.    
\end{displaymath}

An apparent problem with this approach lies in the $ W_f, H_f \in R^{ |V| \times |V| } $ matrices. As the vocabulary can often be in the tens of thousands, this highly increases both the memory and the time complexity of the overall model. However, this problem can be solved by using a dimensionality bottleneck:

\begin{displaymath}
    f(l, t) = H_f ReLU(W_f^T l), \ W_f, H_f \in R^{|V| \times k},
\end{displaymath}

where $ k $ is the dimensionality of the bottleneck and is usually in the low hundreds.

In the simplest form of $ f $, $ t $ can be ignored. This is the same as stating that the gradient does not depend on it and we have a constant vector field as we move through time. Other possibilities are to concatenate it to the input or use it in a conjunction with Hypernetworks \citep{ha2016hypernetworks} to generate $ W_f $ and $ H_f $ based on $ t $. Both approaches are suggested in \citet{chen2018neural} and \citet{grathwohl2018ffjord}. Finally, we can use \emph{Cross-Entropy} for training.

The main idea behind this approach is to solve the Softmax Bottleneck by applying non-linear transformations on the logits, similarly to what is done in \citep{ganea2019breaking}. The difference between the two approaches lies in the nature of the non-linearities applied. \citet{ganea2019breaking} apply monotonic pointwise non-linearities, and here we use a Neural ODE.

\chapter{Continuous Normalizing Flows}
\label{chapter:cnf}

\section{Introduction to Normalizing Flows}
\label{section:cnf:normalizing_flows}

Section \ref{section:limitations:single_transformation} ends with a question: "What if we can start with a simple distribution and distort it into a more complex one?". To answer this question, let us first examine what happens to the densities as we transform some random variable.

Let $ x \in R^d $ be a random variable with an underlying probability density function $ P_X(x) $ and $ f: R^d \mapsto R^d $ be an invertible transformation. Then, if

\begin{displaymath}
    y = f(x),
\end{displaymath}

i.e. we obtain the random variable $ y $ by transforming $ x $ using $ f $, the probability density function $ P_Y(y) $ can be obtained using the change of variables formula

\begin{equation}
    \label{equation:cnf:nf:change_density}
    P_Y(y) = P_X(x) \left | \det \frac{\partial f}{\partial x} \right |^{-1}
\end{equation}

and the change in log density becomes

\begin{equation}
    \label{equation:cnf:nf:change_log_density}
    \log P_Y(y) = \log P_X(x) - \log \left | \det \frac{\partial f}{\partial x} \right |.
\end{equation}

Now let us assume that instead of a single transformation, we want to apply a series of transformations. Let $ f_i: R^d \mapsto R^d, \ i \in \{1, \ ..., \ n\} $ be $ n $ different transformations, and let $ z_0 $ be an initial random variable with a probability density function $ P_{Z_0} $. Then, we can denote the composition of functions $ f_i $ on $ z_0 $ as

\begin{displaymath}
    z_n = f_n \circ ... \circ f_1(z_0),
\end{displaymath}

with the probability density function of $ z_n $ being

\begin{equation}
    \label{equation:cnf:nf:total_change_density}
    P_{Z_n}(z_n) = P_{Z_0}(z_0) \prod_{i=1}^n \left | \det \frac{\partial f_i}{\partial z_{i-1}} \right |^{-1}
\end{equation}

and the total change in log density being

\begin{equation}
    \label{equation:cnf:nf:total_change_log_density}
    \log P_{z_n}(z_n) = \log P_{z_0}(z_0) - \sum_{i=1}^n \log \left | \det \frac{\partial f_i}{\partial z_{i-1}} \right |.
\end{equation}

This technique is called a normalizing flow and was formalized by \citet{rezende2015variational}. They start with a simple probability density function and transform it into a more complex one, by applying a sequence of invertible transformations until a desired level of complexity is obtained. Some simple normalizing flows introduced in their paper \citep{rezende2015variational} are the \emph{planar} and the \emph{radial} flow. The transformation for the planar flow is

\begin{displaymath}
    f(z) = z + uh(w^Tz + b),
\end{displaymath}

where $ u, w \in R^d, b \in R $ and $ h $ is a smooth element-wise non-linearity. On the other hand, the transformation of the radial flow is

\begin{displaymath}
    f(z) = z + \beta h(\alpha, r)(z - z_0),
\end{displaymath}

where $ z_0 \in R^d, \ \alpha \in R^+, \ \beta \in R, \ r = \lvert z - z_0 \rvert $ and $ h(\alpha, \ r) = \nicefrac{1}{\alpha + r} $. The planar flow introduces hyperplanes into the space, and the radial flow introduces spheres into the space.

\begin{figure}[ht]
      \centering
      \includegraphics[width=\columnwidth]{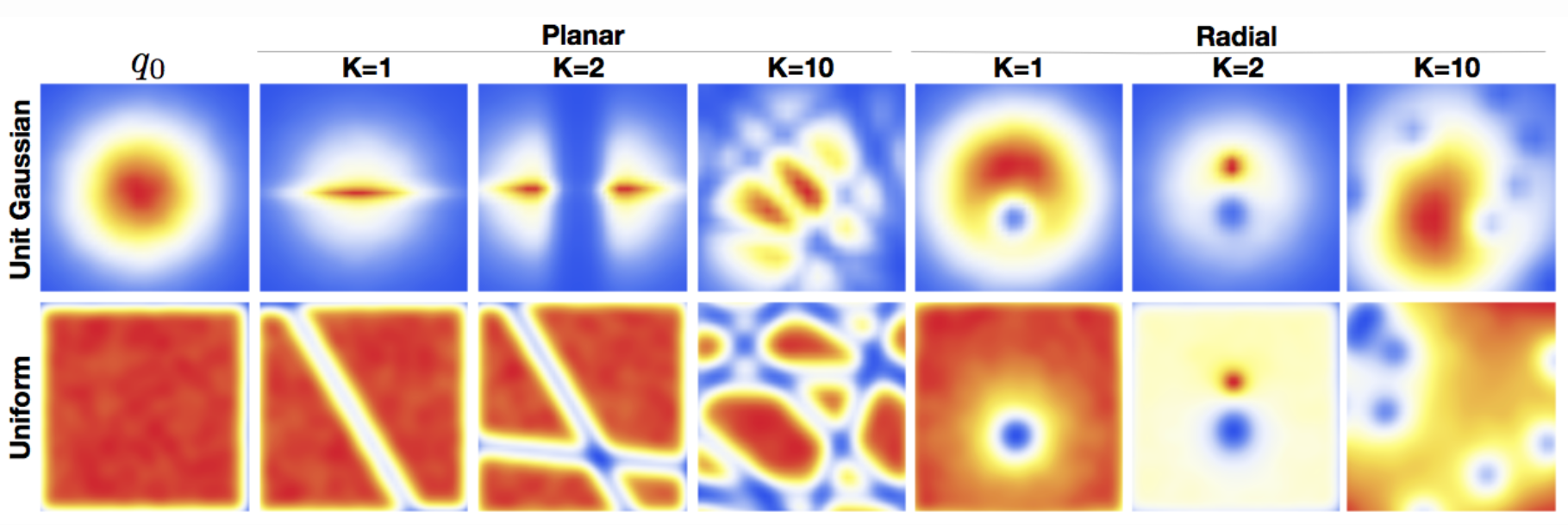}
      \caption{The effect of planar and radial flows on uniform and unit Gaussian distributions \citep{rezende2015variational}.}
      \label{figure:cnf:planar_radial_flows}
\end{figure}

However, in practice, normalizing flows are limited due to their high computational complexity. By looking at equations \ref{equation:cnf:nf:change_density} up to \ref{equation:cnf:nf:total_change_log_density}, we can deduce that normalizing flows require calculating a determinant, which is generally a $ \mathcal{O}(d^3) $ operation. Therefore, their expressiveness is limited by the need to use relatively simple transformations, which Jacobians are easy to compute. For example, both the planar and radial flow allow for linear cost determinant computation and their effect are illustrated on figure \ref{figure:cnf:planar_radial_flows}.

\section{Continuous Normalizing Flows}
\label{section:cnf:cnf}

The previous chapter introduces a novel family of neural models under the name Neural ODEs. \citet{chen2018neural} noticed that the discretized equation \ref{equation:ode:resnet} also appears in Normalizing Flows \citep{rezende2015variational} and the NICE framework \citep{dinh2014nice}. They further realized that performing continuous transformations has an unexpected side effect to the change of variables formula.

\begin{figure}[ht]
      \centering
      \includegraphics[width=0.5\columnwidth]{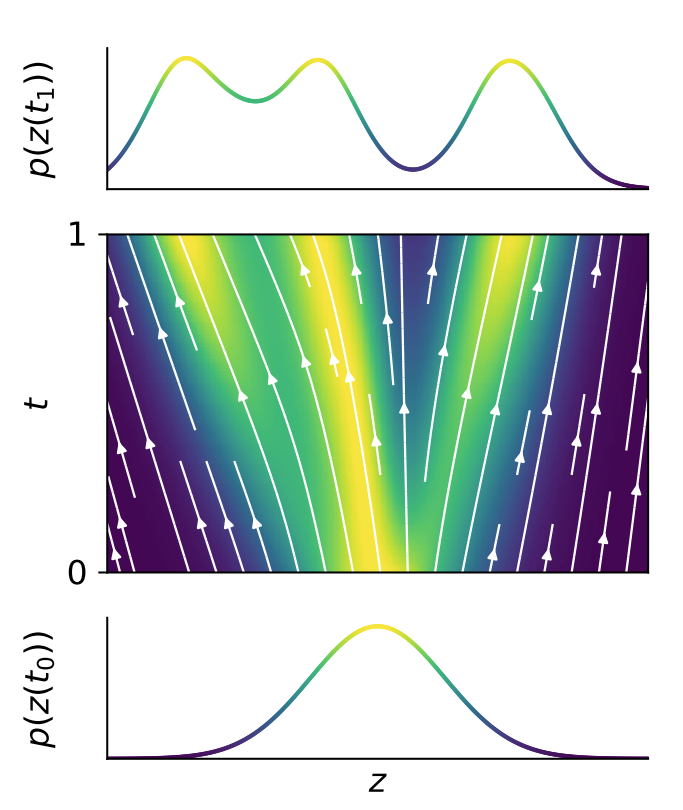}
      \caption{Continuous Normalizing Flows distorting a single mode Gaussian into a more complex distribution \citep{grathwohl2018ffjord}}
      \label{figure:cnf:cnf_transformations}
\end{figure}

Recall from equation \ref{equation:cnf:nf:change_log_density}, that when applying discrete transformations, the change in log density is given by

\begin{align}
    z_{1} &= f(z_{0}), \\
    \log p(z_{1}) &= \log p(z_{0}) - \log \left \lvert \det \frac{\partial f}{\partial z_{0}} \right \rvert,
\end{align}

however, when going into the continuous domain, the change in log density becomes

\begin{align}
    \frac{\partial z(t)}{\partial t} &= f(z(t), t), \\
    \frac {\partial \log p(z(t))} {\partial t} &= -Tr \left( \frac{\partial f}{\partial z(t)} \right),
\end{align}

with the total change in log density given by

\begin{displaymath}
    \log p(z(t_1)) = \log p(z(t_0)) - \int_{t_0}^{t_1} Tr \left( \frac{\partial f}{\partial z(t)} \right) dt.    
\end{displaymath}

\begin{figure}[ht]
      \centering
      \includegraphics[width=\columnwidth]{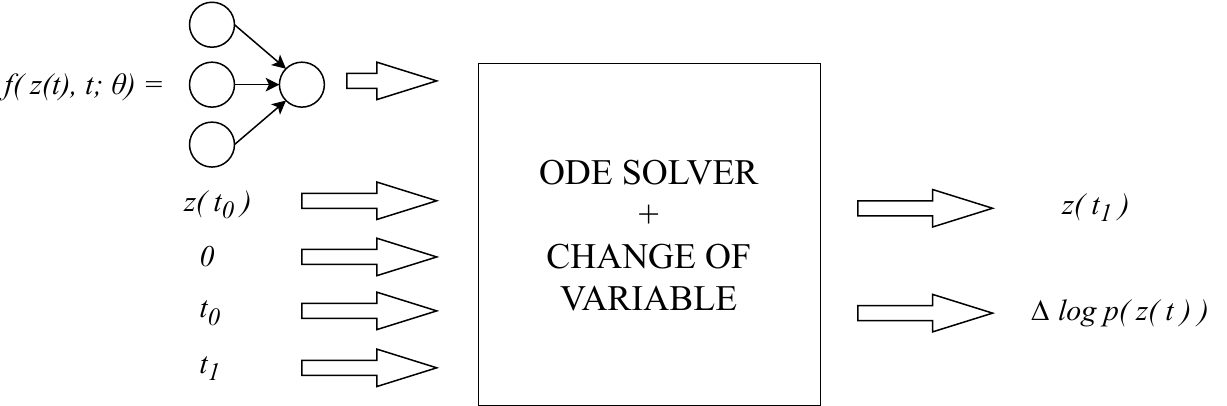}
      \caption{$f(z(t), \ t; \ \theta) $ is neural network specifying the differential equation, $ z(t_0) $ is the initial state, $ 0 $ is the initial log-density, $ t_0 $ is the initial time, $ t_1 $ is the final time, $ z(t_1) $ is the final state and $ \Delta \log p(z(t)) $ is the change in log-density.}
      \label{figure:cnf:cnfs_visualization}
\end{figure}

This combination of NeuralODEs and Normalizing Flows is called \emph{Continuous Normalizing Flows (CNFs)} and can be visualized on Figure \ref{figure:cnf:cnfs_visualization}. One huge difference in the continuous case, is that instead of computing the determinant of the Jacobian, we only need to calculate the trace. Determinants are generally calculated in $ \mathcal{O}(d^3) $, however, the trace is a linear cost operation.

Unfortunately, the overall complexity of the method above is still $ \mathcal{O}(d^2) $ due to the Jacobian, which even though better, is still restrictive. \citet{grathwohl2018ffjord} further optimize the above method and reduce the overall complexity to $ \mathcal{O}(d) $. They achieve this in two steps. First, Vector-Jacobian products can be computed efficiently using reverse-mode automatic differentiation. Secondly, they show that they can get unbiased estimates of the trace of the Jacobian, by using the Hutchinson’s trace estimator \citep{hutchinson1990stochastic}. Consequently, they claim that Continuous Normalizing Flows implemented in this fashion can be considered unrestricted due to the free-form Jacobian of the transformation $ f $.

\citet{grathwohl2018ffjord} additionally provide a PyTorch framework\footnote{\url{https://github.com/rtqichen/ffjord/}} with the previously mentioned improvements. Within their framework we can apply CNFs to random variables and log densities as

\begin{displaymath}
    \underbrace{
        \begin{matrix}
            \begin{bmatrix}
                z(t_1) \\
                \Delta \log p(z_t) 
            \end{bmatrix}
        \end{matrix}
    }_{Solution}
    =
        \bigintss_{t_0}^{t_1}
        \underbrace{
            \begin{matrix}
                \begin{bmatrix}
                    f(z(t), t; \theta) \\
                    Tr \left ( \frac{\partial f}{\partial z(t)} \right)
                \end{bmatrix}
            \end{matrix}
    }_{Dynamics}
    dt = cnf
    \Bigg (
        \underbrace{
            \begin{matrix}
                \begin{bmatrix}
                    z(t_0) \\
                    0
                \end{bmatrix}
            \end{matrix},
        }_{Initial Values}
        t_0, \ t_1, \ f
    \Bigg )
\end{displaymath}

and finally we can obtain $ \log p(z_{t_1}) $  as 

\begin{displaymath}
    \log p(z_{t_1}) = \log p(z_{t_0}) - \Delta \log p(z_t).
\end{displaymath}

\chapter{CNFs for Language Modelling}
\label{chapter:cnf_lm}

\section{Introduction}
\label{section:cnf_lm:introduction}

It was previously shown how Normalizing Flows are a powerful technique to distort simple distributions into complex ones. Until recently, this was only feasible on small toy datasets, however, due to the advances in \citet{chen2018neural} and \citet{grathwohl2018ffjord} and moving to the continuous domain, it has become a viable technique for solving more complex problems.

This chapter goes through several different ways of incorporating Continuous Normalizing Flows in language modelling and discusses both their theoretical and practical implications.

\section{Regression Over Word Embeddings}
\label{section:cnf_lm:regression}

One possible approach is to take the final hidden state $ h $ from an RNN, and use it to parameterize a simple initial distribution $ P_h^0 $, and sample $ z_0 $ from it. Then, we can use CNFs to perform transformations on both $ z_0 $ and its log density in parallel. The goal is to transform $ z_0 $ into the embedding of the word we are trying to predict.

Let $ V $ be the vocabulary, $ h $ be the final hidden state, $ d $ be both the hidden state and the embedding dimensionality and $ e_w^* \in R^{d} $ be the word embedding of the ground truth word. Let $ P_h^0( \ . \ ) $ be the probability density function of the initial distribution parameterized by the hidden state and $ f $ be a neural network architecture parameterizing the gradient as described in chapters \ref{chapter:ode} and \ref{chapter:cnf}. Then:

\begin{align}
    \label{equation:cnf_lm:regression:regression_word_embedding}
    \begin{split}
        z_{t_0} &\sim P_0 \\
        \log p(z_{t_0}) &= \log P(z_{t_0}; h) \\
        z_{t_1}, \Delta \log p (z_t) &= cnf(z_{t_0}, \ \vec{0}, \ t_0 , \ t_1 , \ f) \\
        \log p (z_{t_1}) &= \log p(z_{t_0}) -  \Delta \log  p (z_t)
    \end{split}
\end{align}

We can then treat the problem as a regression problem over word embeddings, i.e. we try to obtain a $ z_{t_1} $ as close as possible to $ e_w^* $ based on some distance metric. We can achieve that by assuming some distribution on the error and then train by minimizing the negative log-likelihood. One possible choice for such a distribution is the Von Mises-Fisher as proposed by \citet{kumar2018mises}.

Let $ e_w^* $ be the ground truth embedding and $ e_w $ be the predicted embedding. The density of the predicted embedding given the ground truth embedding is given by

\begin{displaymath}
    p(e_w; e_w^*) = vMF(e_w; e_w^*) = C_m(\| e_w^* \|) e^{e_w^{*T} e_w},
\end{displaymath}

where $ C_m $ is the normalization constant and is defined as

\begin{displaymath}
    C_m(k) = \frac{ k^{m/2 - 1} } { (2\pi)^{m/2} I_{m/2 - 1}(k) }.
\end{displaymath}

$ I_v $ is a Bessel function of the first kind of order $ v $ and $ m $ is the dimensionality of the embeddings. The Negative Log-Likelihood then becomes

\begin{displaymath}
    NLLvMF(e_w; e_w^*) = - \log C_m(|| e_w ||) - e_w^T e_w^*.
\end{displaymath}

Consequently, by taking equation \ref{equation:cnf_lm:regression:regression_word_embedding} into account the loss becomes

\begin{displaymath}
    loss =  NLLvMF(z_1; \ e_w^*).
\end{displaymath}

Even though we do not explicitly use $ \log p(z_{t_1}) $ for training, at the end of this process we end up with a continuous probability distribution over the entire embedding space. To obtain a distribution over the vocabulary, we have to transform this continuous distribution into a discrete one. Additionally, if word embeddings are not fixed, we would have to add negative samples to prevent clustering them together. 

One drawback of this approach is that this model is not explicitly trained to minimize perplexity. In this thesis, the focus is on models that explicitly minimize perplexity, which is why this type of models was not analyzed further.

\section{CNF Language Models}
\label{section:cnf_lm:cnfs_ce}

\subsection{Training LMs with Cross-Entropy}

The de facto metric for evaluating language models is perplexity, which is why the majority of them are trained using Cross-Entropy. Recall that the Cross-Entropy between two distributions $ p $ and $ q $ is defined as

\begin{displaymath}
    CE(p, \ q) = - \sum_x p(x) \log_b q(x)
\end{displaymath}

and can be interpreted as the average amount of bits needed to encode the outcome of the distribution $ p $ based on a scheme optimized for distribution $ q $. Perplexity on the other hand is defined as

\begin{displaymath}
    PPL(p, \ q) = b^{CE(p, \ q)} = b^{- \sum_x p(x) \log_b q(x)},
\end{displaymath}

where typical choices for the base $ b $ are $ 2 $ or $ e $, simply because logarithms with these bases are easy to compute. It makes no difference which one we go for as long we are consistent across both formulas. In language modeling we usually take $ p $ to be the true distribution and $ q $ to be model distribution obtained from a Softmax layer. As $ p(x) $ is a one-hot encoded ground truth vector, this boils down to

\begin{equation}
    \label{equation:cnf_lm:cnfs_ce:ce_nll}
    CE(p, \ q) = - \log q(x^*),
\end{equation}
    
where $ x^* $ is the ground truth word. From equation \ref{equation:cnf_lm:cnfs_ce:ce_nll} we can deduce that in language modeling, minimizing Cross-Entropy is equivalent to minimizing the negative log-likelihood of $ q $. Additionally, as $ p $ and $ q $ are distributions over words, we are interested in the averaged per-word perplexity. This is a measure of how good our model is, because a per-word perplexity value of $ k $, means that our model's predictive power is just as good as guessing the next word randomly between $ k $ words. Therefore perplexity is considered to be the best metric for evaluating LMs.

\subsection{CNFs for Language Modelling}

Chapter \ref{chapter:limitations} introduced two limitations on standard language models. Let us first introduce how can one apply CNFs to language modeling and what limitations and benefits we get from using them.

Let $ V $ be the vocabulary, $ h \in R^d $ be the final hidden state, $ \log P_h^0(\ . \ ) $ be the initial log-densities commonly obtained from a linear layer with weights $ E \in R^{|V| \times d} $ and additional biases. Finally, let $ f $ be a neural network parameterizing the gradient. Then, the forward pass for one training sample would look like:

\begin{align}
    \label{equation:cnf_lm:cnf_lm_1}
    z_{t_0} &= E , \ z_{t_0} \in R^{|V| \times d} \\
    \label{equation:cnf_lm:cnf_lm_2}
    \log pz_{t_0} &= \log P_h^0(z_{t_0}), \ \log pz_{t_0} \in R^{|V|} \\
    \label{equation:cnf_lm:cnf_lm_3}
    z_{t_1}, \Delta \log pz_t &= cnf(z_{t_0}, \ \vec{0}, \ t_0 , \ t_1 , \ f) \\
    \label{equation:cnf_lm:cnf_lm_4}
    \log pz_{t_1} &= \log pz_{t_0} -  \Delta \log  pz_t
\end{align}

Let us first clarify what \ref{equation:cnf_lm:cnf_lm_1} and \ref{equation:cnf_lm:cnf_lm_2} mean. $ E $ are the weights of the final linear layer used to obtain the logits in standard LMs. A linear layer performs a matrix-vector product between the hidden state $ h $ and the weights $ E $ and adds biases to obtain a resulting vector of size equal to the vocabulary. The components of this vector will correspond to a shifted dot product between $ h $ and the corresponding row in $ E $. The rows of $ E $ are commonly known as \emph{output word embeddings}, contrary to the \emph{input word embeddings} of the initial embedding layer. Additionally, if the weights between the input and the output layers are tied \citep{inan2016tying}, $ E $ is also the embedding matrix. This means that given a hidden state, we obtain a distribution over the word embedding space, with the logits being the corresponding log-densities for each word. Therefore, we can treat the rows of $ E $ as the discrete set of values from this distribution, with log-densities equal to the logits.

Once we have an initial distribution, in equation \ref{equation:cnf_lm:cnf_lm_3} we obtain the change in log-density using CNFs with initial values $ z_{t_0} \in R^{|V| \times d}$ and $ \vec{0} \in R^{|V|} $. Furthermore, $ f $ is an arbitrary neural network and one possible way to define it is

\begin{displaymath}
    \frac{\partial z(t)}{\partial t} = f(z(t), \ t) = W_2 \ ReLU(W_1 \ [z(t), \ t]^T)
\end{displaymath}

where $ W_1 \in R^{d \ \times \ d+1} $ and $ W_2 \in R^{d \times d} $.

In equation \ref{equation:cnf_lm:cnf_lm_4} we obtain the final log-densities by subtracting the change in log-density from the initial distribution. At the end, we obtain log-densities of a distribution over the embedding space. To transform them to discrete probabilities, we can simply take the softmax:

\begin{displaymath}
    q = Softmax(\log pz_{t_1})    
\end{displaymath}

Finally, we train with Cross-Entropy between the obtained model distribution $ q $ and the true distribution $ p $ represented by a one hot encoded ground truth vector:

\begin{displaymath}
    loss = CrossEntropy(p, q)    
\end{displaymath}

As the CNF does not depend on the hidden state, regardless of how many samples we have in a single batch, we only need to perform one flow to obtain the change in log-density. This drastically reduces both the time and memory complexity, however it does mean that we end up with a transformation that is still limited by the \emph{Softmax Bottleneck} problem. When the CNF does not depend on the hidden state it means that the change in log-density for every word is independent of the context. In return, this means that if we represent our model with a matrix similarly to the one in section \ref{section:limitations:softmax_bottleneck} what we are going to get is

\begin{displaymath}
    \begin{matrix}
        \begin{bmatrix}
            \log P_0(w_1 | c_1) - \Delta \log P(w_1) & \hdots & \log P_0(w_M | c_1) - \Delta \log P(w_M)  \\
            \vdots & \ddots & \vdots \\
            \log P_0(w_1 | c_N) - \Delta \log P(w_1) & \hdots & \log P_0(w_M | c_N) - \Delta \log P(w_M)
        \end{bmatrix}
    \end{matrix}
\end{displaymath}

where $ P_0(w_i | c_j) $ is the initial distribution for the $ i $-th word given the $ j $-th context and corresponds to $ \log pz_{t_0} $ in equation \ref{equation:cnf_lm:cnf_lm_4}. Additionally, $ \Delta \log P(w_i) $ is the change in log-density and corresponds to $ \Delta \log pz_t $ in equation \ref{equation:cnf_lm:cnf_lm_4}. We can then split this matrix into two separate matrices $ A $ 

\begin{displaymath}
    \begin{matrix}
    A = \begin{bmatrix}
       \log P_0(w_1 | c_1) & \hdots & \log P_0(w_M | c_1) \\
       \vdots & \ddots & \vdots \\
       \log P_0(w_1 | c_N) & \hdots & \log P_0(w_M | c_N)
      \end{bmatrix}
    \end{matrix}
\end{displaymath}

and $ B $

\begin{displaymath}
    \begin{matrix}
    B = \begin{bmatrix}
       \Delta \log P(w_1) & \hdots & \Delta \log P(w_M) \\
       \vdots & \ddots & \vdots \\
       \Delta \log P(w_1) & \hdots & \Delta \log P(w_M)
      \end{bmatrix}
    \end{matrix}.
\end{displaymath}

Now the original matrix can be written as

\begin{displaymath}
    A - B.
\end{displaymath}

We know that the rank of the sum of two matrices is bounded by

\begin{displaymath}
    \rank(A + B) \leq \rank(A) + \rank(B).
\end{displaymath}

As it was previously shown that $ A $ is a low rank matrix and $ \rank(B) = 1 $, we can conclude that this approach indeed does not solve the \emph{Softmax Bottleneck} problem. Additionally, this model is also restricted by the \emph{Single Transformation} problem described in section \ref{section:limitations:single_transformation}. In the next section, it is discussed how certain changes can release the model from both limitations.

\section{Context Conditioned CNFs}
\label{section:cnf_lm:context_cnfs}

Chapter \ref{chapter:limitations} discusses the current limitations on language modeling. Moreover, in the previous section a novel approach that uses CNFs was introduced that still is still bounded by these limitations. In this section we go through a possible way to upgrade the previously described method in a way that it is not bounded by both the \emph{Softmax Bottleneck} and the \emph{Single Transformation} problems. Namely, the later denotes that we would like to adjust the nature of the distribution based on the context. We can do this by making $ f $ depend on $ h $, i.e. $ f $ becomes

\begin{displaymath}
    f(z(t), \ t, \ h)
\end{displaymath}

and one way to define it is

\begin{equation}
    \label{equation:cnf:context_cnfs:f}
    f(z(t), \ t, \ h) = W_2 \ ReLU(W_1 \ [z(t), \ t]^T + W_h h),
\end{equation}

where $ W_1, W_h \in R^{d \ \times \ d+1} $ and $ W_2 \in R^{d \times d} $.

Let $ V $ be the vocabulary, $ h \in R^d $ be the final hidden state, $ \log P_h^0(\ . \ ) $ be the initial log-densities commonly obtained from a linear layer with weights $ E \in R^{|V| \times d} $ (if the weights are tied \citep{inan2016tying} this is also the embedding matrix) and additional biases. Finally, let $ f $ be a neural network parameterizing the gradient as defined in equation \ref{equation:cnf:context_cnfs:f}. Then, the batched version of the forward pass for one training sample looks like

\begin{align}
    \label{equation:cnf_lm:cnfh_lm_1}
    z_{t_0} &= E , \ z_{t_0} \in R^{|V| \times d} \\
    \label{equation:cnf_lm:cnfh_lm_2}
    \log pz_{t_0} &= \log P_h^0(z_{t_0}), \ \log pz_{t_0} \in R^{|V|} \\
    \label{equation:cnf_lm:cnfh_lm_3}
    z_{t_1}, \Delta \log pz_t &= cnf(z_{t_0}, \ \vec{0}, \ h, \ t_0 , \ t_1 , \ f) \\
    \label{equation:cnf_lm:cnfh_lm_4}
    \log pz_{t_1} &= \log pz_{t_0} -  \Delta \log  pz_t
\end{align}

Even though the approach is same as in section \ref{section:cnf_lm:cnfs_ce}, here the CNF depends on $ h $. We can then proceed and obtain the model distribution $ q $ with

\begin{displaymath}
    q = Softmax(log pz_{t_1})
\end{displaymath}

Finally, we can train the model with Cross-Entropy

\begin{displaymath}
    loss = CrossEntropy(p, q)
\end{displaymath}

where $ p $ is the true distribution and is represented by a one-hot encoded ground truth vector.

To see whether this model still suffers from the \emph{Softmax Bottleneck} problem, we can do the same test as in the previous section. Let us first create a matrix for our model in the same manner

\begin{displaymath}
    \begin{matrix}
    \begin{bmatrix}
       \log P_0(w_1 | c_1) - \Delta \log P(w_1| c_1) & \hdots & \log P_0(w_M | c_1) - \Delta \log P(w_M | c_1)  \\
       \vdots & \ddots & \vdots \\
       \log P_0(w_1 | c_N) - \Delta \log P(w_1| c_N) & \hdots & \log P_0(w_M | c_N) - \Delta \log P(w_M | c_N)
      \end{bmatrix}
    \end{matrix}
\end{displaymath}

where $ P_0(w_i | c_j) $ is the same initial distribution as in the previous case and corresponds to $ \log pz_{t_0} $ in equation \ref{equation:cnf_lm:cnfh_lm_4}. Additionally, $ \Delta \log P(w_i | c_j) $ is the change in log-density and corresponds to $ \Delta \log pz_t $ in equation \ref{equation:cnf_lm:cnfh_lm_4}. Notice that in this case, the change in log-density depends on the context. Then, we can split this into two matrices, $ A $

\begin{displaymath}
    \begin{matrix}
    A = \begin{bmatrix}
       \log P_0(w_1 | c_1) & \hdots & \log P_0(w_M | c_1) \\
       \vdots & \ddots & \vdots \\
       \log P_0(w_1 | c_N) & \hdots & \log P_0(w_M | c_N)
      \end{bmatrix}
    \end{matrix}
\end{displaymath}

and $ B $

\begin{displaymath}
    \begin{matrix}
    B = \begin{bmatrix}
       \Delta \log P(w_1 | c_1) & \hdots & \Delta \log P(w_M | c_1) \\
       \vdots & \ddots & \vdots \\
       \Delta \log P(w_1 | c_N) & \hdots & \Delta \log P(w_M | c_N)
      \end{bmatrix}
    \end{matrix}.
\end{displaymath}

The rank of the original matrix is still bounded by the same rule, i.e.

\begin{displaymath}
    \rank(A + B) \leq \rank(A) + \rank(B)
\end{displaymath}

however, the rank of matrix $ B $ can now go up to $ N $ or $ M $, i.e. $ \rank(B) \leq min(N, \ M) $. Therefore, the rank of $ A + B $ is not constrained by the rank of $ A $ and the sum can potentially be a full-rank matrix. This proves that the model is indeed not limited by the \emph{Softmax Bottleneck}. Additionally, this model starts with a simple initial distribution and distorts it into an arbitrary complex one based on the context. This means that the model is not limited by the \emph{Single Transformation} problem by construction, as we adapt the nature of the distribution based on the context. Therefore, the entire model is more powerful and unrestricted. However, this slight change has certain implications on the computational efficiency of the entire method.

\section{Issues with Context Conditioned CNFs}
\label{section:cnf_lm:issues_cc_cnfs}

Having context conditioned CNFs drastically increases both the time and the memory complexity. Contrary to the previous case where regardless of how many samples in a batch we have solving one CNF was enough, now we need to solve a separate CNF for every distribution. During training, samples are typically represented as tensors of shape

\begin{displaymath}
    [batch\_size, \ sequence\_size, \ embedding\_size],
\end{displaymath}

where $ sequence\_size $ stands for the amount of words in a single sample and $ batch\_size $ stands for the amount of samples in a single batch. As for every sample we need to obtain $ sequence\_size $ distributions, after we process the input with an LM we get a tensor of shape

\begin{displaymath}
    [batch\_size, \ sequence\_size, \ vocabulary\_size],
\end{displaymath}

which contains $ batch\_size \times sequence\_size $ distributions over the vocabulary. This tensor in fact represents the initial distributions for the entire batch. For every distribution we now need to solve a CNF and distort it. If we batch the entire process and solve one large CNF, the tensor $ z_{t_0} $ from equation \ref{equation:cnf_lm:cnfh_lm_1} has shape

\begin{displaymath}
    [batch\_size, \ sequence\_size, \ vocabulary\_size, \ embedding\_size].
\end{displaymath}

The size of the vocabulary in word-based LMs typically starts around $ 10^4 $ and can go up to $ 10^6 $ in datasets like the \emph{The One Billion Word Benchmark} \citep{chelba2013one}. The dimensionality of the embeddings is typically in the low hundreds and reducing it past a certain point results in a loss of performance. Moreover, reducing $ batch\_size $ and $ sequence\_size $ generally only results in a memory versus speed trade-off. This means that evaluating the normalization constant is the real bottleneck for this model.

Possible way to avoid this vocabulary bottleneck is to use character-based LMs. Character based LMs model distributions over sequences of characters, i.e.

\begin{displaymath}
    P(c_1, ..., c_n) = \prod_i^n P(c_i | c_{1..i-1}),
\end{displaymath}

where $ c_i $ is the $ i $-th character in the sequence. Regardless of how many words there are in the training corpus, the vocabulary of character-based LMs is in most cases less than 100. Therefore, the vocabulary is never an issue for character based LMs. However modeling distributions over characters is more complicated than modeling a distribution over words. This follows from the fact that a character can exist in more contexts compared to a word, so modelling all of them is more complicated than in the case of word-based LMs.

However, if we really want to use word based LMs we would have to either resort to techniques like Hierarchical Softmax \citep{morin2005hierarchical} or approximate the normalization constant using sampling techniques. Hierarchical Softmax comes with additional problems, such as choosing the ordering of the words, which is why in this thesis the focus is on approximating the Softmax via sampling.

\section{Softmax Approximations}
\label{section:cnf_lm:softmax_approximations}
\subsection{Introduction to Softmax Approximation}

As shown in the previous section, evaluating the normalization constant for distributions over large vocabularies is not always possible. However, training large vocabulary LMs is still an active research area. \emph{The One Billion Word Benchmark} \citep{chelba2013one} is a large vocabulary benchmark dataset where researchers push the limits and effectively train large vocabulary LMs. In this section we go through approaches that estimate the constant with only a few words.

\subsection{Sampling Based Approaches}

Equation \ref{equation:cnf_lm:cnfs_ce:ce_nll} shows depicts a connection between Cross-Entropy and Negative Log-Likelihood. The Negative Log-Likelihood for a single word and context pair in language modelling can be written as

\begin{displaymath}
    L_{w, c} = - \log \frac{\exp(l_{w,c})}{\sum_{w_i \in V} \exp(l_{w_i,c})},
\end{displaymath}

where $ l_{w,c} $ represents the logit for word $ w $ given context $ c $. If we expand we get

\begin{displaymath}
    L_{w, c} = - l_{w, c} + \log \sum_{w_i \in V} \exp(l_{w_i,c}).
\end{displaymath}

Then, if we take the gradient with respect to the model's parameters $ \theta $ we obtain

\begin{align}
    \nabla_\theta L_{w, c} &= - \nabla_\theta l_{w, c} + \nabla_\theta \log \sum_{w_i \in V} \exp(l_{w_i,c}) \\
                           &= - \nabla_\theta l_{w, c} + \frac{1}{\sum_{w_i \in V} \exp(l_{w_i,c})} \sum_{w_i \in V} \nabla_\theta \exp(l_{w_i,c}) \\
                           &= - \nabla_\theta l_{w, c} + \frac{1}{\sum_{w_i \in V} \exp(l_{w_i,c})} \sum_{w_i \in V} \exp(l_{w_i,c}) \nabla_\theta l_{w_i, c} \\
                           &= - \nabla_\theta l_{w, c} + \sum_{w_i \in V} \frac{\exp(l_{w_i,c})}{\sum_{w_j \in V} \exp(l_{w_j,c})} \nabla_\theta l_{w_i, c}
\end{align}

The fraction inside the sum is the model's probability of $ w_i $ given context $ c $

\begin{displaymath}
    P_\theta(w_i | c) = \frac{\exp(l_{w_i,c})}{\sum_{w_j \in V} \exp(l_{w_j,c})},
\end{displaymath}

so the gradient can be written as

\begin{align}
    \label{equation:cnf_lm:sftmx_aproximations:expectation_1}
    \nabla_\theta L_{w, c} &= - \nabla_\theta l_{w, c} + \sum_{w_i \in V}  P_\theta(w_i | c) \nabla_\theta l_{w_i, c} \\
                           \label{equation:cnf_lm:sftmx_aproximations:expectation_2}
                           &= - \nabla_\theta l_{w, c} + \mathbb{E}_{w_i \sim P_\theta} [ \nabla_\theta l_{w_i, c} ]
\end{align}

Sampling based approaches estimate the expectation in equation \ref{equation:cnf_lm:sftmx_aproximations:expectation_2}.

\subsection{Importance Sampling}
We can approximate the expected value $ \mathbb{E} $ of any probability distribution using Monte-Carlo methods. In our case, we can approximate the expected value in equation \ref{equation:cnf_lm:sftmx_aproximations:expectation_2} with

\begin{displaymath}
    \mathbb{E}_{w_i \sim P_\theta} \nabla_\theta l_{w_i, c} \approx \frac{1}{m} \sum_{i=1}^m \nabla_\theta l_{w_i, c},
\end{displaymath}

where we obtain $ m $ samples from $ P_\theta( w | c ) $ and average the gradients. Unfortunately, $ P_\theta(w | c) $ is what we are trying to learn so we have to use a propositional, or also called a noise, distribution $ Q(w) $ from which it is cheap to sample as a substitute. Furthermore, it is important that $Q$ is similar to $P$, which is why it is taken to be the unigram distribution of the training set in the case of langauge modelling. The goal of \emph{Importance Sampling} is exactly that, to approximate a target distribution $ P $ using a propositional distribution $ Q $ and Monte-Carlo methods. Namely, instead of weighting the gradient in equation \ref{equation:cnf_lm:sftmx_aproximations:expectation_1} with the expensive to compute $ P_\theta( w | c ) $, we weight it with a factor that depends on $ Q(w) $.

A standard practice is to obtain $ k $ noise samples from $ Q $ and estimate the aforementioned quantity with them. Noise Contrastive Estimation (NCE) \citep{gutmann2010noise} proposes using a surrogate loss and it models the problem as a binary classification task, where the goal is to predict whether a sample comes from the true or the noise distribution. \citet{gutmann2010noise} show that as we increase the number of samples, the derivative of the NCE loss approaches the derivative of the softmax function.

\citet{jozefowicz2016exploring} however, propose a different approach based on Importance Sampling. If we take the ground truth word and $ k $ noise samples and define

\begin{displaymath}
    S = \{w_1, ..., w_{k+1} \},
\end{displaymath}

such that $ w_1 $ is always the true word and $ w_2, ... w_{k+1} $ are the noise samples we can optimize a multi-class loss over a multinomial variable $ Y $ representing the labels. Namely, we can define

\begin{displaymath}
    P(Y=k \ | \ S, \ c) = Softmax(l_{w_k, c} - \log Q(w_k))
\end{displaymath}

and train by maximizing the log-likelihood $ \log P(Y = 1 \ | \ S, \ c) $. After training, $ Softmax(l_{w, c}) $ is a good approximation of $ P_\theta(w|c) $. \citet{jozefowicz2016exploring} suggest that this probably is a better choice than NCE for language modeling, as it optimizes a mutli-class classification task instead of a binary one.

\chapter{Experiments}

\section{Legend}
This section explains what the abbreviations in the tables below mean.

\begin{itemize}
    \item \emph{model} - the model being used. AWD stands for AWD-LSTM \citep{merity2017regularizing}, MoS \citep{yang2017breaking} stands for Mixture of Softmaxes and DoC stands for Direct Output Connections \citep{takase2018direct}.
    \item \emph{exp} - number of experts. Models that perform a mixture of distributions need a prespecified value for the number of components in the mixture.
    \item \emph{h} - dimensionality of the middle hidden states of the RNN.
    \item \emph{lasth} - dimensionality of the final hidden state of the RNN.
    \item \emph{emb} - dimensionality of the embeddings.
    \item \emph{lr} - learning rate.
    \item \emph{ep} - epoch at which the presented results are obtained.
    \item \emph{vloss / tloss} - validation loss / test loss. Loss obtained on the validation or the test set.
    \item \emph{vppl / tppl} - validation perplexity / test perplexity. Perplexity obtained on the validation or the test set.
    \item \emph{vbpc / tbpc} - validation bits per character / test bits per character. Bits per character obtained on the validation or the test set.
    \item \emph{prefinetuned} - in the case of transfer learning specifies whether the base model was finetuned before transferring the weights.
    \item \emph{freeze} - in the case of transfer learning specifies whether the transfered weights are fixed or trainable.
\end{itemize}

\section{Datasets}
All models are evaluated on the Penn Treebank dataset which is the standard dataset for evaluating language models. The dataset is used as preprocessed by \citet{mikolov2011empirical} and it consists of 929k training words, 73k validation words, and 82k test words. After preprocessing, all words consist of only lowercase letters and all numbers are replaced with a placeholder $ N $. Additionally, newlines are replaced with a special \textless eos\textgreater \ token. Finally, after preprocessing the dataset contains only the 10k most frequent words and the rest are replaced with a special \textless unk\textgreater \ token.

\section{Word Based Models}

\subsection{Baselines}

The following three models were used as baselines

\begin{enumerate}
    \item AWD-LSTM \citep{merity2017regularizing}
    \item MoS \citep{yang2017breaking}
    \item DOC \citep{takase2018direct}
\end{enumerate}

For every baseline, the latest hyperparameters proposed in their corresponding github repositories were used. Unfortunately, due to changes in PyTorch versions, exact reproduction of their results was not possible. The results before finetuning are presented in table \ref{table:experiments:baselines_word} and the results after finetuning are presented in table \ref{table:experiments:baselines_word_finetuned}.

\begin{table}
\caption{Results from baseline word-based models before finetuning.}
\begin{tabular}{|l|l|l|l|l|l|l|l|l|l|l|}
\hline
\textbf{model}    & \textbf{exp} & \textbf{h}   & \textbf{lasth} & \textbf{emb} & \textbf{lr} & \textbf{ep}  & \textbf{vloss} & \textbf{vppl}  & \textbf{tloss} & \textbf{tppl}  \\ \hline
AWD      & n/a & 960 & 400   & 400 & 20 & 517 & 4.11  & 60.93 & 4.07  & 58.67 \\ \hline
MoS      & 15  & 960 & 620   & 280 & 20 & 511 & 4.06  & 57.89 & 4.02  & 55.84 \\ \hline
DOC      & 15  & 960 & 620   & 280 & 20 & 500 & 4.02  & 55.45 & 3.98  & 53.44 \\ \hline
\end{tabular}
\label{table:experiments:baselines_word}
\end{table}

\begin{table}
\centering
\caption{Results from baseline word-based models after finetuning.}
\begin{tabular}{|l|l|l|l|l|}
\hline
\textbf{model} & \textbf{vloss} & \textbf{vppl}  & \textbf{tloss} & \textbf{tppl}  \\ \hline
AWD   & 4.10  & 60.33 & 4.06  & 58.05 \\ \hline
MoS   & 4.04  & 56.73 & 4.00  & 54.54 \\ \hline
DOC   & 4.00  & 54.68 & 3.97  & 52.87 \\ \hline
\end{tabular}
\label{table:experiments:baselines_word_finetuned}
\end{table}

\subsection{NeuralODE Logit Transformations}
This model is based on Neural ODEs and is explained in section \ref{section:ode:ode_lms}. It performs nonlinear transformations on the logits by using an ODENet on top of the initial AWD-LSTM model. The architecture of the neural network, that specifies the differential equation, in the experiments is defined as

\begin{displaymath}
    f(l, \ t) = W \ Softplus(W_f^T 
        \begin{bmatrix}
           l \\
           t \\
        \end{bmatrix}),
        \ W_f \in R^{|V|+1 \ \times \ k}, \ W \in R^{|V| \ \times \ k},
\end{displaymath}

where $ l $ are the logits obtained from a pre-trained AWD-LSTM, $ t $ is the time and $ W, \ W_f $ are trainable parameters. Additionally, $ |V| $ is the size of the vocabulary and $ k $ is a dimensionality bottleneck and set to be equal to the dimensionality of the embeddings. The hyperparameters being used are the latest hyperparameters proposed in the official AWD-LSTM repository. The results of the experiments for this model are presented in table \ref{table:experiments:neural_odes}.

\begin{table}
\centering
\caption{Results from performing NeuralODE-based transformations on top of the logits of a pre-trained AWD-LSTM model. Several different experiments manage to improve on the baseline. The most notable is experiment 1, as it improves on the baseline by a whole perplexity point. The perplexities on the validation and test sets for this experiment can be seen in bold. Additionally, for this particular model using a learning rate greater or equal to 0.1 results in instant overfitting.}

\begin{tabular}{|l|l|l|l|l|l|l|l|l|}
\hline
\textbf{\#} & \textbf{prefinetuned} & \textbf{freeze} & \textbf{lr} & \textbf{ep} & \textbf{vloss} & \textbf{vppl} & \textbf{tloss} & \textbf{tppl} \\ \hline
1       & no        & no        & 0.01      & 12        & 4.09      & \textbf{59.94}     & 4.06      & \textbf{57.71} \\ \hline
2       & yes       & no        & 0.01      & 5         & 4.10      & 60.50     & 4.06      & 58.09 \\ \hline
3       & no        & yes       & 0.01      & 4         & 4.11      & 60.73     & 4.07      & 58.68 \\ \hline
4       & yes       & yes       & 0.01      & 4         & 4.10      & 60.56     & 4.06      & 58.25 \\ \hline
5       & no        & no        & 0.1       & 1         & 4.09      & 60.02     & 4.06      & 57.73 \\ \hline
6       & yes       & no        & 0.1       & 1         & 4.11      & 60.79     & 4.06      & 58.25 \\ \hline
7       & no        & yes       & 0.1       & 1         & 4.11      & 60.80     & 4.07      & 58.74 \\ \hline
8       & yes       & yes       & 0.1       & 1         & 4.11      & 60.65     & 4.07      & 58.33 \\ \hline
\end{tabular}
\label{table:experiments:neural_odes}
\end{table}

\subsection{Continuous Normalizing Flows}

\subsubsection{Basic CNFs}

This is the most simple model based on Continuous Normalizing Flows and is explained in section \ref{section:cnf_lm:cnfs_ce}. It is more efficient in terms of speed, however theoretically less expressive than the Context Conditioned CNFs model, as it still suffers from both limitations mentioned in chapter \ref{chapter:limitations}. Nonetheless, using the flexibility of CNFs in addition to the base models, seem to be beneficial as the model manages to improve on some of the baselines.

In the experiments, the neural network's architecture, that specifies the differential equation is defined as

\begin{displaymath}
    f(l, \ t) = W \ Softplus(W_f^T 
        \begin{bmatrix}
           l \\
           t \\
        \end{bmatrix}),
        \ W_f \in R^{d+1 \ \times \ d}, \ W \in R^{d \ \times \ d},
\end{displaymath}

where $ d $ is the dimensionality of the embeddings. The hyperparameters being used are the latest hyperparameters proposed in the corresponding repositories of the baseline models.

The results of using CNFs on top of a pre-trained AWD-LSTM model can be seen in table \ref{table:experiments:awd_cnf}. The results of using CNFs on top of a pre-trained MoS model can be seen in table \ref{table:experiments:mos_cnf}. Finally, the results of using CNFs on top of a pre-trained DoC model can be seen in table \ref{table:experiments:doc_cnf}.

\begin{table}
\centering
\caption{Results from training CNFs on top of a pre-trained AWD-LSTM word-based model. Experiment 3 improves on the baseline by 0.3 perplexity points. The perplexities on the validation and test sets for this experiment, can be seen in bold.}

\begin{tabular}{|l|l|l|l|l|l|l|l|l|}
\hline
\textbf{\#} & \textbf{prefinetuned} & \textbf{freeze} & \textbf{lr} & \textbf{ep} & \textbf{vloss} & \textbf{vppl} & \textbf{tloss} & \textbf{tppl}  \\ \hline
1       & no        & no        & 0.1       & 152       & 4.12      & 61.56     & 4.07      & 58.80 \\ \hline
2       & yes       & no        & 0.1       & 160       & 4.15      & 63.20     & 4.11      & 60.97 \\ \hline
3       & no        & yes       & 0.1       & 80        & 4.11      & \textbf{60.65}     & 4.07      & \textbf{58.53} \\ \hline
4       & yes       & yes       & 0.1       & 44        & 4.10      & 60.51     & 4.06      & 58.21 \\ \hline
5       & no        & no        & 1         & 73        & 410       & 60.58     & 4.06      & 57.88 \\ \hline
6       & yes       & no        & 1         & 43        & 4.11      & 60.98     & 4.07      & 58.28 \\ \hline
7       & no        & yes       & 1         & 27        & 4.11      & 61.06     & 4.07      & 58.75 \\ \hline
8       & yes       & yes       & 1         & 21        & 4.12      & 61.31     & 4.08      & 58.88 \\ \hline
\end{tabular}
\label{table:experiments:awd_cnf}
\end{table}

\begin{table}
\centering
\caption{Results from training CNFs on top of a pre-trained MoS word-based model. Experiments 3, 4 and 8 manage to improve on the baseline. Their perplexities on the test and validation set can be seen in bold. For this particular model, freezing the transferred weights of the MoS model and only training the CNF weights results in lower perplexity values.}

\begin{tabular}{|l|l|l|l|l|l|l|l|l|}
\hline
\textbf{\#} & \textbf{prefinetuned} & \textbf{freeze} & \textbf{lr} & \textbf{ep} & \textbf{vloss} & \textbf{vppl} & \textbf{tloss} & \textbf{tppl} \\ \hline
1       & no        & no        & 0.01      & 61        & 4.07      & 58.62     & 4.03      & 56.45 \\ \hline
2       & yes       & no        & 0.01      & 58        & 4.06      & 57.80     & 4.02      & 55.46 \\ \hline
3       & no        & yes       & 0.01      & 355       & 4.06      & \textbf{57.70}     & 4.02      & \textbf{55.78} \\ \hline
4       & yes       & yes       & 0.01      & 338       & 4.04      & \textbf{56.65}     & 4.00      & \textbf{54.53} \\ \hline
5       & no        & no        & 0.1       & 7         & 4.09      & 59.49     & 4.05      & 57.29 \\ \hline
6       & yes       & no        & 0.1       & 7         & 4.07      & 58.77     & 4.03      & 56.35 \\ \hline
7       & no        & yes       & 0.1       & 62        & 4.05      & 57.56     & 4.02      & 55.68 \\ \hline
8       & yes       & yes       & 0.1       & 62        & 4.03      & \textbf{56.50}     & 4.00      & \textbf{54.42} \\ \hline
\end{tabular}
\label{table:experiments:mos_cnf}
\end{table}

\begin{table}
\centering
\caption{Results from training CNFs on top of a pre-trained DoC word-based model. None of the experiments improve on the baselines. Similarly to previous experiments, smaller learning rates and freezing the transferred weights of the base model achieves lower perplexity values.}
\begin{tabular}{|l|l|l|l|l|l|l|l|l|}
\hline
\textbf{\#} & \textbf{prefinetuned} & \textbf{freeze} & \textbf{lr} & \textbf{ep} & \textbf{vloss} & \textbf{vppl} & \textbf{tloss} & \textbf{tppl} \\ \hline
1       & no           & no     & 0.05 & 11 & 4.04  & 56.69 & 4.00  & 54.54 \\ \hline
2       & yes          & no     & 0.05 & 8  & 4.03  & 56.05 & 3.99  & 54.15 \\ \hline
3       & no           & yes    & 0.05 & 81 & 4.02  & 55.55 & 3.98  & 53.59 \\ \hline
4       & yes          & yes    & 0.05 & 64 & 4.00  & 54.79 & 3.97  & 53.02 \\ \hline
5       & no           & no     & 0.1  & 11 & 4.05  & 57.16 & 4.00  & 54.85 \\ \hline
6       & no           & yes    & 0.1  & 43 & 4.02  & 55.63 & 3.98  & 53.66 \\ \hline
7       & yes          & yes    & 0.1  & 46 & 4.00  & 54.86 & 3.97  & 53.14 \\ \hline
8       & no           & no     & 1    & 49 & 4.06  & 57.84 & 4.01  & 55.14 \\ \hline
9       & no           & yes    & 1    & 11 & 4.07  & 58.62 & 4.04  & 56.58 \\ \hline
10       & yes          & yes    & 1    & 15 & 4.06  & 58.00 & 4.03  & 56.10 \\ \hline
\end{tabular}
\label{table:experiments:doc_cnf}
\end{table}

\subsubsection{Context Conditioned CNFs}

Due to the issues discussed in section \ref{section:cnf_lm:issues_cc_cnfs} all word-based Context Conditioned CNFs are trained using Importance Sampling. In every training iteration, 20 labels are obtained from the unigram distribution of the training set and are concatenated to the true label. This drastically speeds up training, however evaluating on the original validation and test sets remains unfeasible. Therefore, when evaluating Context Conditioned CNFs, only the first 400 samples from the validation and the test sets are used.

The hyperparameters being used are the latest hyperparameters proposed in the official MoS repository. Furthermore, the RNN base of the model is initialized with the weights of a pre-trained AWD-LSTM model. Additionally, the architecture of the neural network, that specifies the differential equation, in the experiments is defined as

\begin{align*}
    f(l, \ t, \ h) &= W \ Softplus(W_f^T 
        \begin{bmatrix}
           l \\
           t \\
        \end{bmatrix} + W_h h) \\
        W_f &\in R^{d+1 \ \times \ d} \\ 
        W, W_h &\in R^{d \ \times \ d},
\end{align*}

where $ l $ are the logits, $ t $ is time, $ h $ is the final hidden state obtained from the AWD-LSTM base and $W, \ W_f $ and $ W_h $ are trainable parameters. The results can be seen in table \ref{table:experiments:cnfh_word}. Additionally, for fair comparison, table \ref{table:experiments:awd_mini_eval} contains metrics for the AWD-LSTM model, computed on the first 400 samples of the validation and test sets.

\begin{table}
\centering
\caption{Results from training Context Conditioned CNFs on top of a pre-trained word-based AWD-LSTM model. The perplexities are computed only on the first 400 samples of the validation and test set. No experiment improves on the baseline in table \ref{table:experiments:awd_mini_eval}. However, the baseline model was trained by evaluating the full partition function of the Softmax, and this model is trained by randomly sampling 20 labels from the unigram distribution of the training set in every iteration. Due to the small number of labels in every iteration and the randomness, a lot more epochs are needed to reach or improve on the performance of the baseline model.}

\begin{tabular}{|l|l|l|l|l|l|l|l|l|}
\hline
\textbf{\#} & \textbf{prefinetuned} & \textbf{freeze} & \textbf{lr} & \textbf{ep} & \textbf{vloss} & \textbf{vppl} & \textbf{tloss} & \textbf{tppl} \\ \hline
1       & no        & no        & 0.01      & 84        & 4.12      & 61.45     & 3.93      & 51.09 \\ \hline
2       & yes       & no        & 0.01      & 84        & 4.10      & 60.52     & 3.96      & 52.48 \\ \hline
3       & no        & yes       & 0.01      & 46        & 4.12      & 61.72     & 3.94      & 51.35 \\ \hline
4       & yes       & yes       & 0.01      & 46        & 4.10      & 60.14     & 3.96      & 52.58 \\ \hline
5       & no        & no        & 0.1       & 52        & 4.11      & 61.15     & 3.94      & 51.51 \\ \hline
6       & yes       & no        & 0.1       & 45        & 4.10      & 60.64     & 3.96      & 52.48 \\ \hline
7       & no        & yes       & 0.1       & 24        & 4.13      & 62.22     & 3.95      & 52.14 \\ \hline
8       & yes       & yes       & 0.1       & 24        & 4.11      & 60.85     & 3.95      & 52.00 \\ \hline
\end{tabular}
\label{table:experiments:cnfh_word}
\end{table}

\begin{table}
\centering
\caption{Perplexities of the baseline word-based AWD-LSTM model computed on the first 400 samples of the validation and test sets.}
\begin{tabular}{|l|l|l|l|l|}
\hline
\textbf{model} & \textbf{vloss} & \textbf{vppl}  & \textbf{tloss} & \textbf{tppl}  \\ \hline
AWD   & 4.09  & 59.64 & 3.92  & 50.30 \\ \hline
\end{tabular}
\label{table:experiments:awd_mini_eval}
\end{table}

\section{Character Based Models}

Training character-based models on Penn Tree Bank \citet{mikolov2011empirical} is performed using same pre-processing as when training word-based models. This means that only the 10000 most frequent words are kept, and all others are substituted with an \textless unk\textgreater \ token. This drastically reduces the number of possible transitions between characters and simplifies the problem.

Additionally, when using Context Conditioned CNFs for character models the size of the vocabulary is usually less than 50. Therefore, character based Context Conditioned CNFs do not suffer from the issues mentioned in section \ref{section:cnf_lm:issues_cc_cnfs} and are trained using the entire vocabulary.

For character-based models only the AWD-LSTM \citep{merity2017regularizing} mode was used as a baseline. Similarly to the word-based models, the latest hyperparameters proposed in the official github repository were used. Exact reproduction of the results was not possible due to differences in PyTorch versions. The baseline results can be seen in table \ref{table:experiments:baselines_char}.

Evaluating the full partition function, even though feasible, is still not fast enough to train these models from scratch. Therefore, similarly to the case of word-based models, the RNN weights are initialized with the weights of the pre-trained baseline AWD-LSTM model. The results can be seen in table \ref{table:experiments:cnfh_characters}.

\begin{table}
\centering
\caption{Results from training baseline character-based models.}
\begin{tabular}{|l|l|l|l|l|l|l|l|l|l|}
\hline
\textbf{model} & \textbf{h} & \textbf{lasth} & \textbf{emb} & \textbf{lr} & \textbf{ep} & \textbf{vloss} & \textbf{vbpc} & \textbf{tloss} & \textbf{tbpc} \\ \hline
awd       & 1000       & 200            & 200          & 0.002       & 364         & 0.84           & 1.212         & 0.82           & 1.183         \\ \hline
\end{tabular}
\label{table:experiments:baselines_char}
\end{table}

\begin{table}
\centering
\caption{Results from training Context Conditioned CNFs on top of a pre-trained character-based AWD-LSTM model. Most of the experiments reach same performance as the baseline model. An issue with training Context Conditioned CNF's with transferred weights is the possibility of initializing the model in a local minimum. This means that the CNF learns weights almost equal to 0, in order to retrieve the initial distribution from the baseline model.}

\begin{tabular}{|l|l|l|l|l|l|l|l|}
\hline
\textbf{\#} & \textbf{freeze} & \textbf{lr} & \textbf{ep} & \textbf{vloss} & \textbf{vbpc} & \textbf{tloss} & \textbf{tbpc} \\ \hline
1       & yes             & 1e-5        & 55          & 0.84           & 1.213         & 0.82           & 1.184         \\ \hline
2       & no              & 1e-4        & 15          & 0.84           & 1.212         & 0.82           & 1.182         \\ \hline
3       & yes             & 1e-4        & 28          & 0.84           & 1.212         & 0.82           & 1.183         \\ \hline
4       & no              & 2e-3        & 1           & 0.92           & 1.329         & 0.90           & 1.295         \\ \hline
5       & yes             & 2e-3        & 3           & 0.84           & 1.212         & 0.82           & 1.183         \\ \hline
\end{tabular}
\label{table:experiments:cnfh_characters}
\end{table}

\chapter{Conclusion}

Language modelling remains one of the most important subfields of Natural Language Processing. Additionally, its real life applications are increasing everyday. As such, it is an extremely active research area of extreme importance to the Natural Language Processing community. Especially important class of models are the RNN-based language models. Due to the relatively small number of parameters in comparison to transformer-based models, they represent the de facto model for commercial language modelling. However, recent findings \citep{yang2017breaking} show several limitations on standard RNN-based language modelling.

In this thesis, I have introduced several different models for language modelling based on a novel concept. Namely, I have successfully integrated NeuralODEs \citep{chen2018neural} and Continuous Normalizing Flows \citep{grathwohl2018ffjord} with RNN-based language models. For every model, I have introduced and discussed the theoretical and practical limitations. Finally, I have summarized my findings and I have managed to improve on some of the baselines.

\section{Future Work}

The main limitation of this novel family of language models is their computational complexity. Since fast and efficient training was not possible, there was no room for extensive hyperparameter search. This means that most of the models manage to either beat their respective baseline or get close to it, by using hyperparameters that are not optimized for them. As performing an extensive hyperparameter search usually results in an improvement of several metric points, it is interesting to see how much of an improvement we are going to see.

For the same reason, all models are trained by first initializing them with the weights of the baseline model they are compared against. It is possible that this initialization puts the model in a sub-optimal region of the optimizational space, as seems to be the case with character-based models. Given enough resources and time, it is interesting to see how will these models perform when trained from scratch.

\appendix

\chapter{Abbreviations}

\begin{table}[h]
    \centering
    \begin{tabular}{r l}
        NLP & Natural Language Processing \\
        LM & Language Modelling \\
        RNN & Reccurent Neural Network \\
        LSTM & Long Short-Term Memory \\
        MoS & Mixture of Softmaxes \\
        DOC & Direct Output Connections \\
        SGD & Stochastic Gradient Descent \\
        ASGD & Averaged Stochastic Gradient Descent \\
        NT-ASGD & Non-monotonically Triggered ASGD \\
        ResNet & Residual Network \\
        ODE & Ordinary Differential Equation \\
        CNF & Continuous Normalizing Flow \\
        NCE & Noise Contrastive Estimation \\
    \end{tabular}
\end{table}

\backmatter

\bibliography{refs}


\end{document}